\documentclass[twoside]{article}

\usepackage[accepted]{aistats2023}
\usepackage[round]{natbib}
\renewcommand{\bibname}{References}
\renewcommand{\bibsection}{\subsubsection*{\bibname}}

\usepackage{microtype}
\usepackage{graphicx}
\usepackage{booktabs} 

\usepackage{hyperref}

\usepackage{amsmath}
\usepackage{amssymb}
\usepackage{mathtools}
\usepackage{amsthm}

\usepackage[capitalize,noabbrev]{cleveref}

\theoremstyle{plain}

\theoremstyle{definition}

\theoremstyle{remark}

\usepackage{amsmath,amsfonts,bm}

\def\eqref#1{equation~\ref{#1}}

\def\1{\bm{1}}

\def\mH{{\bm{H}}}

\def\mM{{\bm{M}}}

\def\mU{{\bm{U}}}
\def\mV{{\bm{V}}}
\def\mW{{\bm{W}}}

\def\mY{{\bm{Y}}}

\DeclareMathAlphabet{\mathsfit}{\encodingdefault}{\sfdefault}{m}{sl}
\SetMathAlphabet{\mathsfit}{bold}{\encodingdefault}{\sfdefault}{bx}{n}

\def\gG{{\mathcal{G}}}

\usepackage[T1]{fontenc}
\usepackage{mysymbols}
\usepackage{booktabs} 
\usepackage{amsmath}
\usepackage{amsthm}
\usepackage{amssymb}
\usepackage{stmaryrd}
\usepackage{latexsym}
\usepackage{bm}
\usepackage{subcaption}
\usepackage{placeins}
\theoremstyle{plain}
\usepackage{mathtools}
\newcommand{\defeq}{\vcentcolon=}
\usepackage{thmtools,thm-restate}
\usepackage{algorithm}
\usepackage{algorithmic}
\usepackage{cleveref}
\usepackage{wrapfig}
\usepackage{multirow}
\newcommand{\vhalf}{\vspace{-.05in}}

\usepackage{enumitem}
\usepackage{setspace}
\usepackage{multicol}

\begin{document}

%

%
\runningauthor{Yang, Dzanic, Petersen, Kudo, Mittal, Tomov, Camier, Zhao, Zha, Kolev, Anderson, Faissol}

\twocolumn[

\aistatstitle{Reinforcement Learning for Adaptive Mesh Refinement}

\aistatsauthor{ Jiachen Yang$^*$ \And Tarik Dzanic$^*$ \And Brenden Petersen$^*$ \And Jun Kudo$^*$ \And Ketan Mittal  }

\aistatsaddress{ LLNL \And  Texas A\&M University \And LLNL \And LLNL \And LLNL } 

\aistatsauthor{ Vladimir Tomov \And Jean-Sylvain Camier \And Tuo Zhao \And Hongyuan Zha \And Tzanio Kolev }

\aistatsaddress{ LLNL \And LLNL \And Georgia Tech \And Georgia Tech \And LLNL } 

\aistatsauthor{ Robert Anderson \And Daniel Faissol }

\aistatsaddress{ LLNL \And LLNL } 

]


\begin{abstract}
\vspace{-15pt}
Finite element simulations of physical systems governed by partial differential equations (PDE) crucially depend on adaptive mesh refinement (AMR) to allocate computational budget to regions where higher resolution is required. Existing scalable AMR methods make heuristic refinement decisions based on instantaneous error estimation and thus do not aim for long-term optimality over an entire simulation. We propose a novel formulation of AMR as a Markov decision process and apply deep reinforcement learning (RL) to train refinement {\it policies} directly from simulation. AMR poses a challenge for RL as both the state dimension and available action set changes at every step, which we solve by proposing new policy architectures with differing generality and inductive bias. The model sizes of these policy architectures are independent of the mesh size and hence can be deployed on larger simulations than those used at training time. We demonstrate in comprehensive experiments on static function estimation and time-dependent equations that RL policies can be trained on problems without using ground truth solutions, are competitive with a widely-used error estimator, and generalize to larger and unseen test problems.
\end{abstract}

\section{INTRODUCTION}
\label{sec:intro}

Numerical simulation of PDEs via the finite element method (FEM) \citep{brenner2007mathematical} plays an integral role in computational science and engineering \citep{reddy2010finite,monk2003finite}.
Given a fixed set of basis functions, the resolution of the finite element mesh determines the trade-off between solution accuracy and computational cost. 
For problems with large variations in local solution characteristics, uniform meshes can be computationally inefficient due to their suboptimal distribution of mesh density, under-resolving regions with complex features such as discontinuities or large gradients and over-resolving regions with smoothly varying solutions.
For systems with multi-scale properties, 
attempting to resolve these features with uniform meshes can be challenging even on the largest supercomputers.
To achieve more efficient numerical simulations, adaptive mesh refinement (AMR), a class of methods that dynamically adjust the mesh resolution during a simulation to maintain equidistribution of error, is used to significantly increase accuracy relative to computational cost.


Existing methods for AMR share the iterative process of computing a solution on the current mesh, estimating refinement indicators, marking element(s) to refine, and generating a new mesh by refining marked elements \citep{bangerth2013adaptive, Cerveny2019}.
The optimal algorithms for error estimation and marking in many problems, especially evolutionary PDEs, are not known \citep{bohn2021recurrent}, and deriving them is difficult for complex refinement schemes such as $hp$-refinement \citep{zienkiewicz1989effective}.
As such, the current state-of-the-art is guided largely by heuristic principles that are derived by intuition and expert knowledge \citep{zienkiewicz1992superconvergent}, such as using an instantaneous error estimator with greedy element marking, but choosing the best combination of heuristics is complex and not well understood.
Whether and how optimal AMR strategies can be found by directly optimizing a long-term performance objective are open questions.

\begin{figure}[t]
    \centering
    \includegraphics[width=1.0\linewidth]{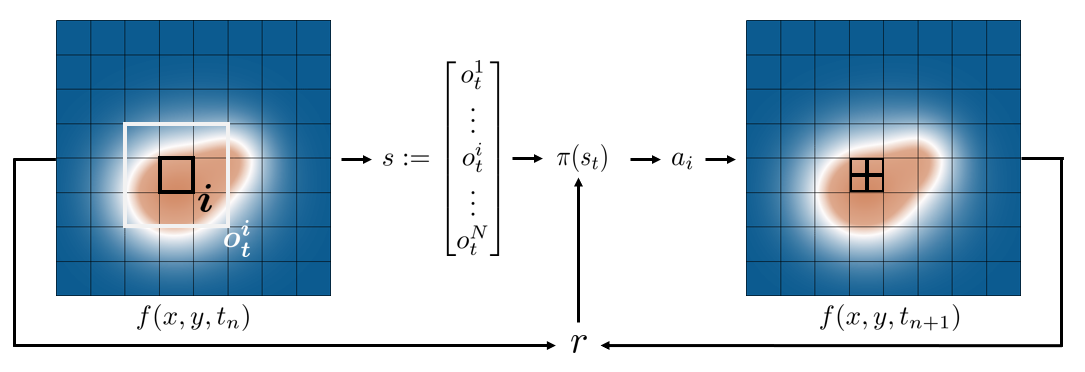}
    \caption{AMR viewed as a Markov decision process.}
    \label{fig:rl_loop}
\end{figure}

We advance the novel notion that adaptive mesh refinement is fundamentally a {\it sequential decision-making} problem: making an optimal sequence of refinement decisions to optimize cumulative or terminal accuracy, subject to a computational budget.
We hypothesize that a sequence of greedy decisions based on instantaneous error indicators does not constitute an optimal sequence of decisions.
This is because the optimality of a current refinement decision depends on complex interactions between solution dynamics, budget consumption, and the propagation of numerical error from the current decision throughout the system, all of which impact solution accuracy many steps into the future and are not revealed by instantaneous error. 
In time-dependent problems for example, an error estimator by itself cannot preemptively refine elements which would encounter complex features in the next time step.


Given this perspective, we formulate AMR as a Markov decision process (MDP) \citep{puterman2014markov} (\Cref{fig:rl_loop}) and propose a reinforcement learning (RL) \citep{sutton2018reinforcement} approach to train a mesh refinement policy to optimize a performance metric, such as final solution error.
AMR poses a challenge for RL as the sizes of the state and set of available actions depend on the current number of mesh elements, which changes with each refinement action at every MDP time step.
One may define a fixed bounded state and action space given a finite refinement budget, but this is inefficient as the policy's input-output dimensions must accommodate the full exponentially large space---e.g., input dimensions on the order of millions of degrees of freedom in large applications---whereas only subspaces (with increasing size) are encountered.
This motivates us to design efficient policy architectures that leverage the known correspondence between each mesh state and valid actions.


Our paper contributes a proof of feasibility for an entirely novel way of learning AMR strategies using deep RL.
In particular:
1) We formally define an MDP with effective variable-size state and action spaces for AMR (\Cref{subsec:mdp});
2) We propose three policy architectures---with differing generality and inductive bias for modeling interaction---that operate on such variable-size spaces (\Cref{sec:architecture});
3) Toward the eventual goal of deploying on large and complex problems on which RL cannot tractably be trained, we propose to train on small representative features with known analytic solutions and using a novel reward formulation that applies to problems without known solutions (\Cref{sec:experimental-setup});
4) Our experiments demonstrate for the first time that RL can outperform a greedy refinement strategy based on the widely-used Zienkiewicz-Zhu-type error estimator; moreover, we show that an RL refinement policy can generalize to higher refinement budgets and larger meshes, transfer effectively from static to time-dependent problems, and can be effectively trained on more complex problems without readily-available ground truth solutions (\Cref{sec:results}).


\section{RELATED WORK}

To the best of our knowledge, our work is the first to formulate AMR as a global sequential decision-making problem and show the feasibility of an RL approach.
A contemporaneous single-agent local approach centers the decision-making on each individual element \citep{foucart2022deep}, requiring different definitions of the environment transition between training and test time for scalability.
\citet{brevis2020data} apply supervised learning to find an optimal parameterized test space without modifying the degrees of freedom.
\citet{bohn2021recurrent} show theoretically that the estimation and marking steps of AMR for an elliptic PDE can be represented optimally by a recurrent neural network, but
model optimization was not addressed.

Previous work have trained neural networks to predict static (non-adaptive) mesh densities and  sizes for use by downstream mesh generators \citep{dyck1992determining,chedid1996automatic,zhang2020meshingnet,pfaff2020learning,chen2020output}.
More recently, neural network policies have been trained via RL to generate a mesh incrementally from initial boundary vertices \citet{pan2023reinforcement}.
Recent studies have used graph neural networks (GNN) \citep{sperduti1997supervised,gori2005new,scarselli2008graph} to predict PDE dynamics on general unstructured and non-uniform meshes \citep{alet2019graph,belbute2020combining,pfaff2020learning}.
Other work use neural networks as function approximators in numerical solvers to achieve faster convergence, generalization, and higher resolution of coarse simulations \citep{hsieh2018learning,luz2020learning, bar2019learning}.
Our work focuses on optimizing a finite element space via training a policy that changes the mesh, rather than predicting dynamics or learning components of a solver.

Adjoint-based methods for goal-oriented AMR can optimize a cost such as terminal solution accuracy \citep{offermans2017adjoint,rannacher2014model,apel2014graded}, but they incur significant complexity such as a forward-backward solve and a checkpointing system for the backwards-in-time solution \citep{becker2001optimal}, which limits their scalability to coarse grids \citep{davis2020analysis}.


Both state and action space sizes change at every time step within an episode in AMR, whereas
RL has been typically applied to environments with fixed-size observation and small bounded action spaces in almost all benchmark problems \citep{mnih2015human,brockman2016openai,osband2019behaviour}.
Applications where the available action set varies with state \citep{berner2019dota,vinyals2019grandmaster} do not face the challenge of potentially millions of possible actions that arises in large-scale AMR.
Applications of RL to graph-structured problems handle states with increasing size, but the action space does not grow with the size of the graph \citep{you2018graph,trivedi2020graphopt}.

\section{BACKGROUND AND FORMULATION}

\subsection{Finite Element Method}

Our mesh adaptation strategy is implemented in a FEM-based framework
\citep{brenner2007mathematical}.
In FEM, the domain $\Omega \subset \Rbb^D$ is modeled with a mesh that is a union of $E$ nonoverlapping subsets (\emph{elements}) such that $\Omega := \bigcup \Omega_k$ where
$k \in \mathbb{N} : k \leqslant E$ (e.g., see \Cref{fig:rl_loop}).
The solution on these elements is represented using polynomials (\emph{basis functions}) which are used to transform the
governing equations into a system of algebraic equations via the weak formulation.
AMR is a commonly used approach to improve the trade-off between the solution accuracy, which depends on the shape and sizes of elements, and the computational cost, which depends on the number of elements.
The most ubiquitous method for AMR is $h$-refinement, whereby elements are split into smaller elements (refinement) or multiple elements coalesce to form a single element (derefinement). 
\subsection{AMR as a Markov Decision Process}
\label{subsec:mdp}

We formulate AMR with spatial $h$-refinement\footnote{Polynomial $p$-refinement can be formulated in a similar way. $r$-refinement \citep{huang2010adaptive, TMOP18} can be formulated as an RL problem but is not treated in this work.}
as a Markov decision process $\Mcal \defeq (\Ocal, N_{\text{max}}, \Acal, R, P, \gamma)$ with each component defined as follows.
Each episode consists of $T$ RL time steps: for time-dependent PDEs, $T$ spans the entire simulation and there may be multiple underlying PDE evolution steps per RL step; for static problems, $T$ is an arbitrary number of steps at which RL can act.
Consider a time step $t$ when the current mesh has $N_t \leq N_{\text{max}} \in \Nbb$ elements.
Each element $i$ is associated with an \textit{observation} $o^i_t \in \Ocal$ and the \textit{global state} is $s_t \defeq [o^1_t,\dotsc,o^N_t] \in \Ocal^{N_t}$.
We define $\Ocal \defeq \Rbb^d$ such that each element's observation is a tensor of shape $d \defeq l \times w \times c$ that includes the values and refinement depths of a local window centered on itself (see \Cref{app:observation}).
For brevity, let $\Scal_t$ denote the current global state space $\Ocal^{N_t}$.
We denote an action by $a_t \in \Acal_t \defeq \lbrace 0, 1,\dotsc,N_t \rbrace \subset \Acal \defeq \lbrace 0, 1,\dotsc,N_{\text{max}} \rbrace,$ where $0$ means ``do-nothing'' and $i\neq 0$ means refine element $i$.
Given the current state and action, the MDP transition $P$ consists of:
\begin{enumerate}[topsep=0pt, partopsep=0pt, itemsep=0pt, label={}, leftmargin=0cm]
\item 1) refining the selected element into multiple finer elements (which increases $N_t$) if a refinement budget $B$ is not exceeded and the selected element is not at the maximum refinement depth $d_{\text{max}}$;
\item 2) stepping the finite element simulation forward in time (for time-dependent PDEs only);
\item 3) computing a solution on the new finite element space.
\end{enumerate}
Steps 1-3 are standard procedures in FEM \citep{anderson2019mfem}, knowledge of which is not used in our proposed model-free RL approach.
Although the size of the state vector and set of valid actions changes with each time step due to the varying $N_t$, this MDP is well-defined since one can define the global state space as the union of all $\Ocal^N, N < N_{\text{max}}$, and likewise for the action space.
An agent moves through subspaces of increasing size during an episode.

When a true solution is available at {\it training} time, the reward at step $t$ is defined as the change in error from the previous step, normalized by the initial error to reduce variation across function classes:
\vspace{-3pt}
\begin{align}\label{eq:reward}
    r_t \defeq (\lVert e_{t-1} \rVert_2 - \lVert e_t \rVert_2)/\lVert e_0 \rVert_2 \, ,
\end{align}
where error $e$ is computed relative to the true solution. 
With abuse of notation, we shall use $e$ to indicate the error norm.
The ground truth is not needed to deploy a trained policy on test problems.
When the true solution is not readily available, as is the case for most non-trivial PDEs, one may run a reference simulation on a highly-resolved mesh to compute \eqref{eq:reward}, but this approach can be prohibitively expensive for training on large-scale simulations. Instead, we propose the use of 
a \textit{surrogate reward} $r_t \coloneqq \lVert u_{t,\text{refine}} - u_{t,\text{no-refine}} \rVert_2$, the normed difference between the estimated solution $u$ with and without executing the chosen refinement action.
This surrogate, which is an upper bound on the true reward and effectively acts as an estimate of the error reduction, is only used at training time, whereas at test time, the effectiveness of trained policies is evaluated using the error with respect to a highly-resolved reference simulation.

Our objective to find a stochastic policy $\pi \colon \Scal_t \rightarrow \Delta(\Acal_t)$ to maximize the objective
\vspace{-5pt}
\begin{align}\label{eq:objective}
    J(\pi) \defeq \Ebb_{a\sim \pi(\cdot|s), s_{t+1 }\sim P(\cdot|a,s_t)} \left[ \sum_{t=1}^{T} \gamma^t r_t \right] \, .
\end{align}
Aside from $\gamma \in (0,1)$, the dense and shaped reward \citep{ng1999policy} defined in \eqref{eq:reward} implies that maximizing this objective is equivalent to maximizing total error reduction: $e_0 - e_{\text{final}}$.
We work with the class of policy optimization methods as they naturally admit stochastic policies that could benefit AMR at test time: a stochastic refinement action could reveal the need for further refinement in a region that appears flat on a coarse mesh.
We build on REINFORCE and PPO \citep{sutton2000policy,schulman2017proximal} to train a policy $\pi_{\theta}$ (parameterized by $\theta$) using batches of trajectories $\lbrace \tau_k \defeq \lbrace (s_t, a_t, r_t)_k \rbrace_{t=1}^T \rbrace_{k=1}^K$ generated by the current policy.
By virtue of RL, $\pi_{\theta}$ is trained not merely to act greedily but to account for dependence of future rewards on current actions.

\begin{figure*}[t]
    \centering
    \begin{subfigure}[t]{0.49\linewidth}
    \includegraphics[trim=-3cm -3cm 0 0, width=0.95\linewidth]{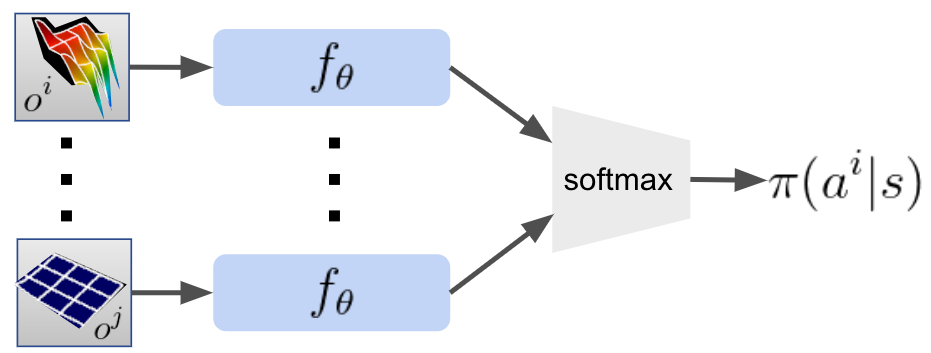}
    \caption{Independent Policy Network}
    \label{fig:ipn}
    \end{subfigure}
    \hfill
    \begin{subfigure}[t]{0.49\linewidth}
    \includegraphics[trim=-5cm 0 0 0, width=0.95\linewidth]{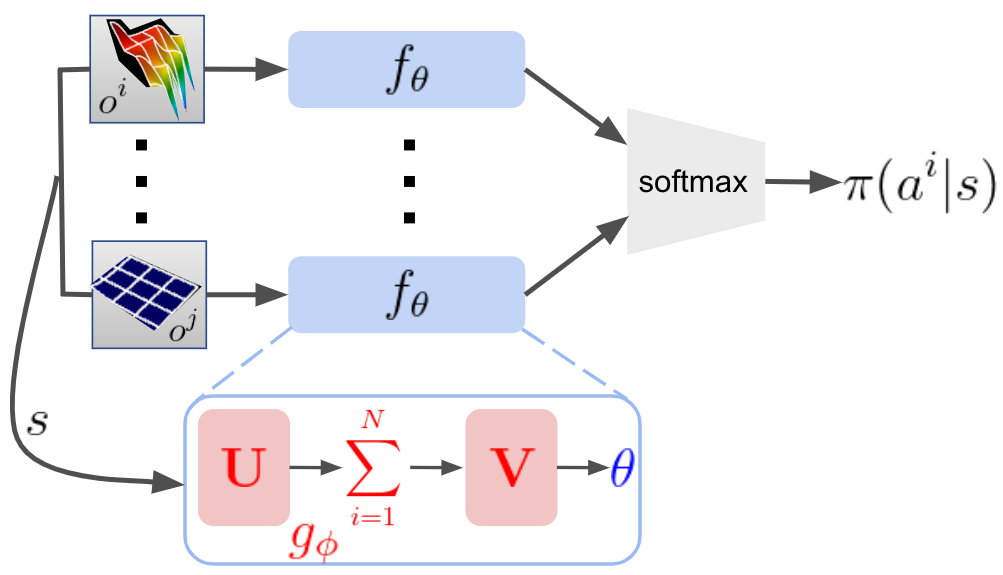}
    \caption{Hypernetwork policy}
    \label{fig:hypernet}
    \end{subfigure}
    \hfill
    \begin{subfigure}[t]{0.8\linewidth}
    \includegraphics[trim=-10cm 0 0 0, width=0.9\linewidth]{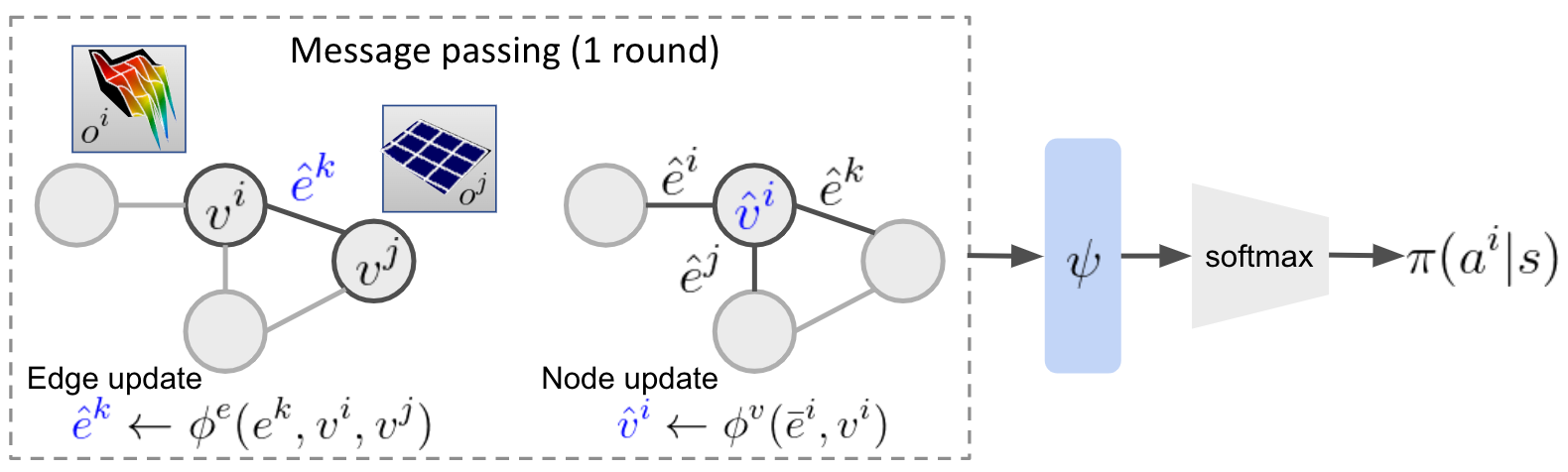}
    \caption{Graph network policy}
    \label{fig:graphnet}
    \end{subfigure}
    \caption{RL policy architectures for adaptive mesh refinement. 
    (a) Independent Policy Network (IPN) applies the same function $f_{\theta}$ to map each element's observation $o^i$ independently to logits, which are then mapped to action probabilities via a global \texttt{softmax}. 
    (b) The Hypernetwork policy generates the parameters $\theta$ of the main network via a hypernetwork $g_{\phi}$ that receives global state $s$.
    (c) The Graph Network policy conducts multiple rounds of message passing, using learned edge network $\phi^e$ and node network $\phi^v$ to update edge and node features individually and in parallel for all edges and nodes. Final node features are mapped to logits, on which a global \texttt{softmax} is applied.}
    \label{fig:architecture}
\end{figure*}

\section{POLICY ARCHITECTURES FOR VARIABLE STATE-ACTION SPACES}
\label{sec:architecture}
\vhalf

The exact $1{:}1$ correspondence between the number of observation components and the number of valid actions calls for dedicated policy architectures for AMR.
Different inductive biases expressed by different model architectures can have significant impact on performance even when used in the same learning algorithm \citep{battaglia2018relational}.
As such, we investigate three policy architectures that address the challenge of variable size state vector $s \in \Rbb^{N_t \times d}$ and action set $\lbrace 0, 1,\dotsc,N_t \rbrace$, where number of elements $N_t$ changes during each episode.
These architectures are compatible with any stochastic policy gradient algorithm.
We focus on the special case of $1{:}1$ correspondence between the number of observations that compose each global state and the number of available actions at that state.
Although not treated in this work, these policy architectures can be easily extended to the general case of $1{:}k$ correspondence\footnote{To include de-refinement actions ($k=2$), the only change is to map each element’s observation to
two output logits, one interpreted as refinement and the other de-refinement, and include both in the global \texttt{softmax} over all elements.}.


\subsection{Independent Policy Network}
\label{subsec:ipn}

The Independent Policy Network (IPN) handles the $1{:}1$ correspondence by mapping each observation to a probability for the corresponding action.
Let $f_{\theta} \colon \Rbb^d \mapsto \Rbb$ be a function parameterized by $\theta$.
Given a matrix of observations $s := [o^1, \dotsc, o^N] \in \Rbb^{N \times d}$, we define the policy as
\begin{align}\label{eq:independent-policy}
    \pi(\cdot|s) = \texttt{softmax}\left( f_{\theta}(o^1), \dotsc, f_{\theta}(o^N) \right) \, .
\end{align}
For example, using a neural network with hidden layer $\mW \in \Rbb^{d \times h}$ with $h$ nodes, output layer $\mH \in \Rbb^{h \times 1}$, and activation function $\sigma$, the discrete probability distribution over $N$ actions conditioned on $s$ is defined by $\texttt{softmax}\left( \sigma(s \mW) \mH \right)$.
IPN is illustrated in \Cref{fig:ipn}.

IPN applies to meshes of variable size since the set of trainable parameters $\theta$ is independent of $N$, but it has two main limitations.
Firstly, it makes a strong assumption of locality as the action probability at an element does not depend on the observations at other elements.
This assumption also appears in existing AMR methods that estimate error independently at each element; in fact, the output probabilities of IPN may be viewed as normalized error estimates.
Secondly, the permutation equivariance of this architecture---i.e., $\pi(a^{\mu(i)}|(o^{\mu(1)},\dotsc,o^{\mu(N)})) = \pi(a^i|s)$ for any permutation operator $\mu \colon [N] \mapsto [N]$---means that 
one cannot use the ordering of inputs to represent spatial relations among elements, which would be necessary for refining an element based on neighboring conditions.
We mitigate this problem by defining each observation as an image that includes neighborhood information and using a convolutional layer, but this may face difficulties on unstructured meshes with non-quadrilateral elements \citep{Cerveny2019}.

\subsection{Hypernetwork Policy}
\label{subsec:hypernet}

The hypernetwork policy captures higher-order interaction among inputs via the function form (illustrated in \Cref{fig:hypernet})
\begin{align}
    \pi(\cdot|s) = \texttt{softmax}\left(f_{g_{\phi}(s)}(o^1),...,f_{g_{\phi}(s)}(o^N)\right).
\end{align}
The main policy network weights $\theta$ are now the output of a hypernetwork \citep{ha2017hypernetworks} $g_{\phi} \colon \Rbb^{N \times d} \mapsto \Rbb^{\text{dim}(\theta)}$, parameterized by $\phi$, which produces mixing among the inputs $s \in \Rbb^{N \times d}$.
Continuing with the example in IPN, where the policy network's first layer is $\mW \in \Rbb^{d \times h}$,
a hypernetwork with two layers can be instantiated as $\left[ \sum_{i=1}^N \left( s \mU \right)_{i,:} \right] \mV = \mW$
where $\mU \in \Rbb^{d \times h_1}$ and $\mV \in \Rbb^{h_1 \times (d \times h)}$
are the trainable parameters $\phi$, and $\mM_{i,:}$ denotes the $i$-th row of matrix $\mM$.
The output $\mW$ is then used as part of $\theta$ in \eqref{eq:independent-policy}.

This increased generality comes with more difficulty in the choice of $g_{\phi}$, which affects the extent to which it captures interaction among inputs.
It does not contain an inductive bias for the local nature of interactions in physical PDEs.
In fact, the use of a summation from $i=1$ to $N$ in the example above means that complete global information affects each local refinement decision, which is a strong inductive bias.

\subsection{Graph Network Policy}
\label{subsec:graphnet}

We build on graph networks \citep{scarselli2008graph,battaglia2018relational} to address both the issue of interaction terms and spatial relation among elements.
Specifically, we construct a policy based on Interaction Networks \citep{battaglia2016interaction}, illustrated in \Cref{fig:graphnet}, which is a special case without global attributes\footnote{
While not demonstrated in this work, one may define a global graph attribute containing the PDE coefficients or initial/boundary conditions and apply variants of Graph Networks that use global attributes in their update rules to improve generalization.}.
At each step, the mesh is represented as a graph $\gG = (V, E)$.
Each vertex $v^i$ in $V = \lbrace v^i \rbrace_{i=1:N}$ corresponds to element $i$ and is initialized to be the observation $o^i$.
$E = \lbrace (e^k, r^k, s^k) \rbrace_{k=1:N^e}$
is a set of edges with attributes $e^k$ between sender vertex $s^k$ and receiver vertex $r^k$.
An edge exists between two vertices if and only if they are spatially adjacent.
We define the initial edge attribute $e^k$ as a one-hot vector indicator of the difference in refinement depth between $r^k$ and $s^k$.

Graph networks capture the relations between nodes and edges via the inductive bias of its update rules.
The model we use consists of a learned edge network $\phi^e$ and a node network $\phi^v$, which are applied individually and in parallel to all edges and nodes.
A single \textit{forward pass} through the graph policy involves multiple rounds of \textit{message passing} (see \Cref{alg:graphnet}).
Each round is defined by the following operations: 1) Each edge attribute $e^k$ is updated by learned function $\varphi^e$ using local node information via $\hat{e}^k \leftarrow \varphi^e(e^k, v^{r^k}, v^{s^k})$;
2) For each node $i$, we denote by $\hat{E}^i \coloneqq \lbrace (\hat{e}^k, r^k, s^k)\rbrace_{r^k=i}$ the set of all edges with node $i$ as the receiver, and all updated edge attributes are aggregated into a single feature $\bar{e}^i \leftarrow \rho^{e \rightarrow v}(\hat{E}^i)$ by aggregation function $\rho^{e \rightarrow v}$ (e.g., element-wise sum);
3) Then, each node attribute is updated by $\hat{v}^i \leftarrow \varphi^v(\bar{e}^i, v^i)$ using learned function $\varphi^v$.
Each round increases the size of the neighborhood that determines node attributes.
Finally, we map each node attribute to a scalar using learned function $\psi$, apply a global softmax over all nodes, and interpret the value at each node $i$ as the probability of choosing element $i$ for refinement.

The graphnet policy addresses both limitations of the IPN and the hypernetwork policy.
Cross terms arise in the forward pass due to mutual updates of edge and node attributes using local information.
The order of cross terms increases with each message-passing round.
Local spatial relations between mesh elements are included by construction in the adjacency matrix and initial edge attributes, so there is no need to include global spatial information in each element's observation vector.

\vhalf
\section{EXPERIMENTAL SETUP}
\vhalf
\label{sec:experimental-setup}

Our experiments assess the ability of RL, using the proposed policy architectures, to find AMR strategies that
generalize to test function\footnote{In the sense of testing a policy's performance after training, not in the sense of the weak formulation of PDEs.} classes that differ from the training class, generalize to variable mesh sizes and refinement budgets, and extend to more complex problems without readily-available ground truth solutions.
We define the FEM environment in \Cref{subsec:femenvironment},
the train-test procedure in \Cref{subsec:procedure}, and the implementation of our method and baselines in \Cref{subsec:implementation}.

\subsection{AMR Environment}
\label{subsec:femenvironment}

\textbf{MFEM.}
We use MFEM \citep{anderson2019mfem, mfem-web}, a modular open-source C++ library for FEM, to implement the MDP for AMR.
We ran experiments on two classes of AMR problems: static and time-dependent. 
In the static case, the objective of mesh refinement is to minimize the $L^2$ error norm of projecting a variety of test functions onto a two-dimensional $H^1$ finite element space.
In the time-dependent case, the functions are projected onto a two-dimensional $L^2$ finite element space, and a PDE of the form $\frac{\partial u}{\partial t} + \nabla {\cdot}\mathbf{F}(u) = 0$
is solved on a periodic domain using the finite element framework.
Unlike the static case, the numerical error accumulated at each time step propagates with the physical dynamics and determines future error.
Two types of PDEs were used: the linear advection equation, where $\mathbf{F}(u) = \mathbf{c}u$ with $\mathbf{c} = [1,0]$, and the nonlinear Burgers equation, where $\mathbf{F}(u) = \mathbf{c}u^2$ with $\mathbf{c} = [1,0.3]$. The advection equation is used as there is an analytic solution to provide a ground truth, whereas the Burgers equation is used as a representative of more complex physical systems, including shock and rarefaction waves, without a readily-available analytic solution. The solution is represented using continuous (or discontinuous) second-order Bernstein polynomials for the static (or time-dependent) case, and the initial mesh is partitioned into $n_x \times n_y$ quadrilateral elements.
\Cref{app:mfemctrl} contains further on the FEM and MDP implementations.

\begin{figure}[t]
\centering
\begin{subfigure}[t]{0.24\linewidth}
    \centering
    \includegraphics[width=1.0\linewidth]{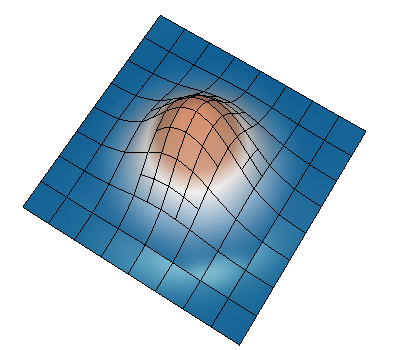}
    \caption{Bumps}
    \label{fig:mesh_bumps_pg}
\end{subfigure}
\begin{subfigure}[t]{0.24\linewidth}
    \centering
    \includegraphics[width=1.0\linewidth]{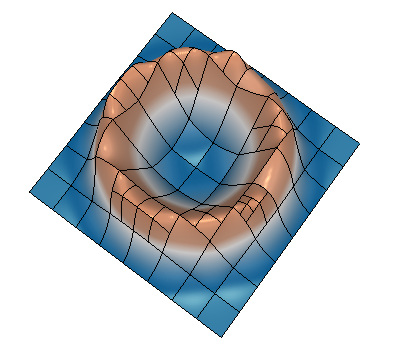}
    \caption{Circles}
    \label{fig:mesh_circles_pg}
\end{subfigure}
\begin{subfigure}[t]{0.24\linewidth}
    \centering
    \includegraphics[width=1.0\linewidth]{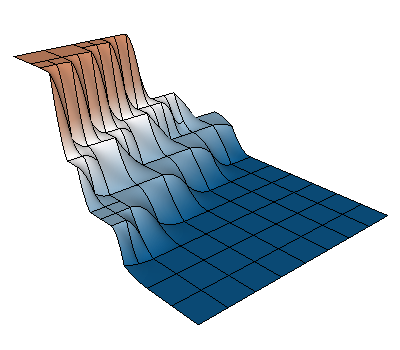}
    \caption{Steps}
    \label{fig:mesh_steps_pg}
\end{subfigure}
\begin{subfigure}[t]{0.24\linewidth}
    \centering
    \includegraphics[width=1.0\linewidth]{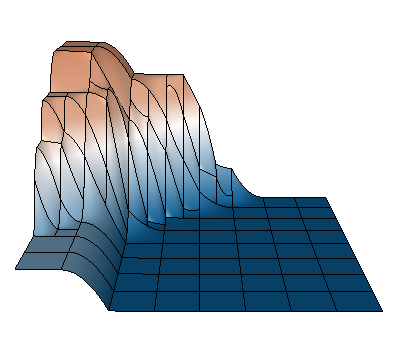}
    \caption{Steps2}
    \label{fig:mesh_steps2_pg}
\end{subfigure}
\caption{Examples from each true solution function class.
}
\label{fig:mesh_static}
\vspace{-10pt}
\end{figure}

\textbf{True solutions.}
We defined a collection of parameterized function classes, each exhibiting features such as sharp discontinuities and smooth variations, from which we randomly sample ground truth functions $f \colon [0,1]^2 \mapsto \Rbb$ to initialize each episode.
The collection, shown in \Cref{fig:mesh_static} and defined precisely in \Cref{app:ground_truth}, includes: \textit{bumps}, \textit{circles}, \textit{steps},
and \textit{steps2} (a combination of two steps).
These functions with closed form allow us to compute the error and reward at train time for static and advection problems.
In the case of Burgers equation where the exact solution is not readily-available, we either use reference simulations on a highly-resolved mesh to act as a ``ground truth'' or employ the surrogate reward (defined in \Cref{subsec:mdp}) to compute the reward for training.

In the static case, the true solution is fixed and each simulation time step is an RL step. 
For the time-dependent PDE cases, the initial solution is transported through the periodic domain and the ratio of simulation time steps to RL steps is set such that a feature advecting at unit velocity returns to its original position after 10 RL steps.
We set episode length at training time to equal the refinement budget $B$, with $B=10$ for static problems, $B=20$ for advection, and $B=50$ for Burgers.
Larger budget (i.e., longer episode) was chosen for the time-dependent problems to test the ability of RL to find long-term refinement strategies that outperform greedy baselines.
All methods were subject to the same budget constraint.
Due to the Gibbs phenomena in FEM, using smooth polynomial approximations to solve hyperbolic systems containing discontinuities can introduce spurious oscillations which, in turn, can cause the simulation to become unstable. Therefore, we limit the true solutions to smooth functions (e.g., \textit{bumps}, \textit{circles}) for the advection and Burgers cases. Due to the nonlinearity of Burgers equation, initially smooth solutions can develop discontinuities in finite time. This behavior is resolved using the flux-corrected transport (FCT) approach \citep{boris1997fct}.

\begin{figure*}[t]
\centering
\begin{subfigure}[t]{0.28\linewidth}
    \centering
    \includegraphics[height=0.68\linewidth]{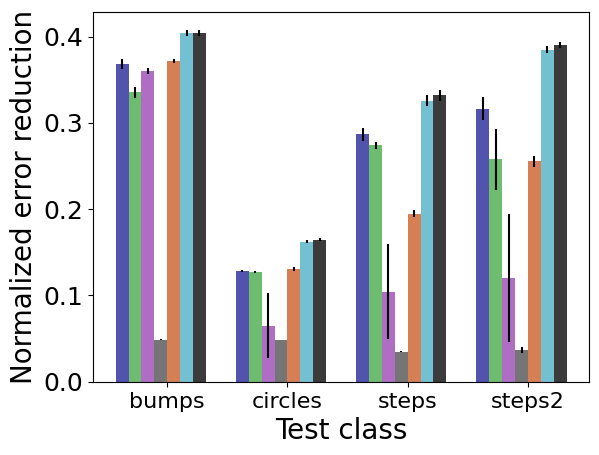}
    \vspace{-.075in}
    \caption{Static}
    \label{fig:static_train_equal_test}
\end{subfigure}
\hfill
\begin{subfigure}[t]{0.28\linewidth}
    \centering
    \includegraphics[height=0.68\linewidth]{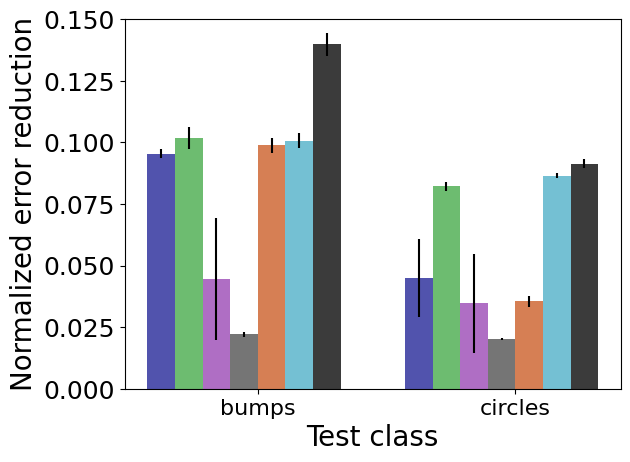}
    \vspace{-.075in}
    \caption{Advection}
    \label{fig:advection_train_equal_test}
\end{subfigure}
\hfill
\begin{subfigure}[t]{0.28\linewidth}
    \centering
    \includegraphics[height=0.68\linewidth]{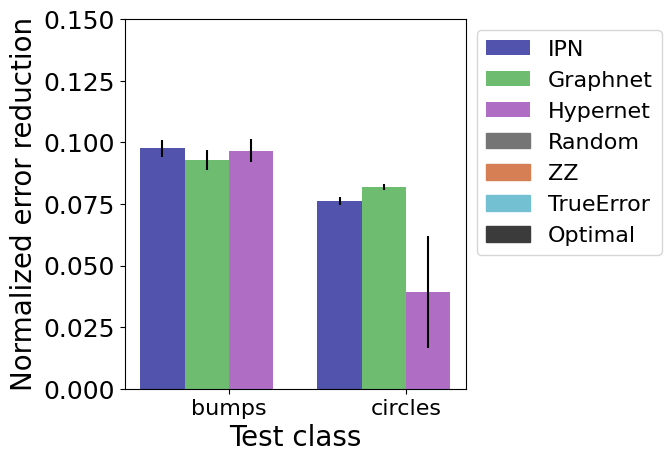}
    \vspace{-.235in}
    \caption{Static$\rightarrow$advection}
    \label{fig:static_to_advection}
\end{subfigure}
\vspace{-.1in}
\caption{\textbf{In-distribution} and \textbf{Static$\rightarrow$advection}. Performance of IPN, Graphnet and Hypernetwork policies versus baselines.
Higher values are better. (a,b) RL policies were trained and tested on the same function class, for static and advection cases independently. (c) Static-trained policies on a function class are tested on advection of the same class.
}
\label{fig:train_equal_test}
\end{figure*}

\subsection{Experiments and Performance Metric}
\label{subsec:procedure}

We conducted the following experiments 
to compare RL policies with baselines:
\begin{enumerate}[topsep=0pt, partopsep=0pt, itemsep=0pt, label={}, leftmargin=0cm]
    \item \textbf{In-distribution}: Train and test on true solutions sampled from the same function class.
    Training and testing on different function classes is unlikely in applications, since expert knowledge of solution features is almost always available and one can train on similar functions, but we include experimental results in \Cref{app:results}.
    \item \textbf{Out-of-distribution:} For problems such as Burgers equation where training on multiple initial conditions (ICs) may be expensive, we show the effectiveness of policies trained on a single initial condition (IC) when tested on multiple random ICs, either with or without fine-tuning.
    \item \textbf{Generalization}: 1) \textbf{Static$\rightarrow$advection}: Policies trained on static functions are tested on advection.
    2) \textbf{Budget$\uparrow$}: Policies trained with a small refinement budget $B$ (20 on static and 10 on advection) are test with $B=50, 100$.
    3) \textbf{Size$\uparrow$}: Policies trained on an $8 \times 8$ mesh are tested on meshes up to size $200 \times 200$ (360k solution nodes), a 625x increase in element number, with and without preserving the relative solution and mesh length scales.
\end{enumerate}

We define the performance of a given refinement policy in an episode in the static case as $(e_{\text{initial}} - e_{\text{final}}) / e_{\text{initial}}$, where $e_{\text{initial}}$ (or $e_{\text{final}}$) is the error norm at the beginning (or end) of an episode, to remove the variation in the error due to different true solution classes and random function initialization within each class.
In the time-dependent case, without any refinement, the error may increase over the course of the simulation due to the accumulation of discretization error.
Hence, we define performance as
$(e_{\text{no-refine, final}} - e_{\text{final}}) /  e_{\text{initial}}$, where $e_{\text{no-refine, final}}$ is the final error without any refinement.

For every experiment and every policy architecture, we trained four independent policies with different random seeds.
For each test case, we report the mean and standard error---over the four independent policies and with different simulator seeds
---of the mean performance metric over 100 test episodes.
At each test episode, all methods faced the same initial condition (which differs across episodes).

\subsection{Implementation and Baselines}
\label{subsec:implementation}

We describe the high-level implementation here and provide complete details in \Cref{app:implementation}.
All policy architectures use a convolutional neural network with the same architecture as the input layer.
The \textbf{IPN} has two fully-connected hidden layers with $h_1$ and $h_2$ nodes and \texttt{ReLU} activation, followed by a \texttt{softmax} output layer.
Its action on input states is described in \Cref{subsec:ipn}.
The \textbf{Graphnet} policy is implemented with the Graph Nets library \citep{battaglia2018relational}. 
Each input state consists of node observation tensors, all edge vectors, and the adjacency matrix.
Node tensors are first passed through an Independent block, after which multiple Interaction networks \citep{battaglia2016interaction} act on both node and edge embeddings to produce a probability at each node (see \Cref{subsec:graphnet}).
The \textbf{Hypernet policy} is parameterized by matrices $\mU \in \Rbb^{d \times h_1}$, $\mV \in \Rbb^{h_1 \times (d \times h)}$, and $\mY \in \Rbb^{d \times h}$, where $h_1$ and $h$ are design choices.
$\mU$ and $\mV$ act on input state $s$ to produce the main policy weights $\mW \in \Rbb^{d \times h}$ while $\mY$ acts on $s$ to produce a bias $b \in \Rbb^h$, so that the main policy's first hidden layer is $\texttt{ReLU}(s\mW + b)$.
Output probabilities are computed in the same way as IPN.

\textbf{Baselines.}
The \textbf{ZZ} policy uses a Zienkiewicz-Zhu-type recovery-based error estimator
\citep{zienkiewicz1992superconvergent} and refines the element with the largest estimated error.
The \textbf{TrueError} policy refines the element where the error of the numerical solution with respect to the true solution is largest.
It is effectively an upper bound on the performance of any state-of-the-art method based on instantaneous error estimator, but it is \textit{not} the theoretical upper bound on performance because refining the element with largest current error does not necessarily result in smallest final error.
It cannot be deployed without known solutions.
The \textbf{GreedyOptimal} policy performs one-step lookahead by checking all possible outcomes of refining each element individually and chooses the element whose refinement would result in the lowest error.
It is intractable even for simple time-dependent PDEs on relatively coarse meshes.
TrueError and GreedyOptimal are strong \textit{oracles} that cannot be used in real applications.

\begin{figure*}[htb!]
\centering
\begin{subfigure}[t]{0.24\linewidth}
    \centering
    \includegraphics[width=1.0\linewidth]{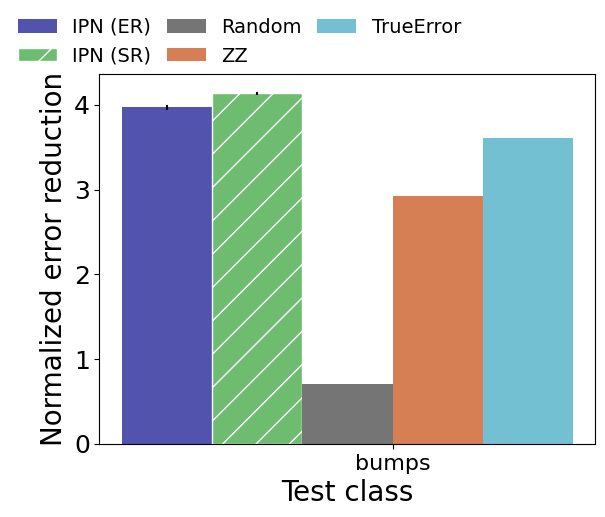}
    \vspace{-.15in}
    \caption{Test on fixed IC}
    \label{fig:burgers_barchart_single_0p9_vertical0p3_surrogate}
\end{subfigure}
\hfill
\begin{subfigure}[t]{0.24\linewidth}
    \centering
    \includegraphics[width=1.0\linewidth]{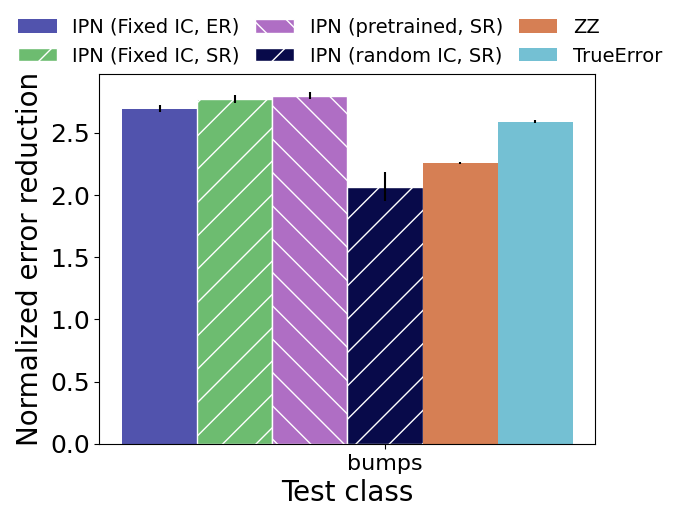}
    \vspace{-.15in}
    \caption{Test on random ICs}
    \label{fig:burgers_barchart_10x10_steps50_randomIC}
\end{subfigure}
\hfill
\begin{subfigure}[t]{0.24\linewidth}
    \centering
    \includegraphics[width=1.0\linewidth]{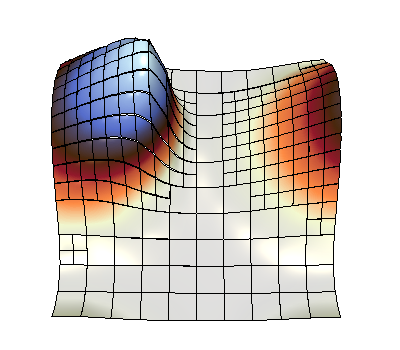}
    \vspace{-.15in}
    \caption{Mesh by IPN (Fixed IC, SR)}
    \label{fig:burgers_single_t50}
\end{subfigure}
\hfill
\begin{subfigure}[t]{0.24\linewidth}
    \centering
    \includegraphics[width=1.0\linewidth]{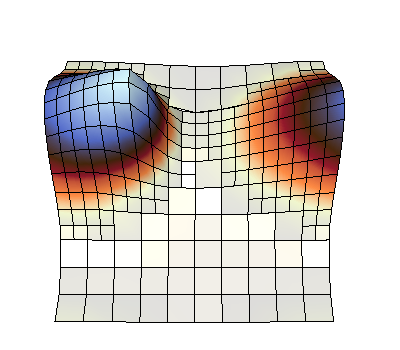}
    \vspace{-.15in}
    \caption{Mesh by IPN (pretrained, SR)}
    \label{fig:burgers_pretrainsingle_trainrandom_randomIC_t50}
\end{subfigure}
\vspace{-5pt}
\caption{\textbf{Burgers equation} and \textbf{surrogate reward}. Solid bars/ER denote exact reward, striped bars/SR denote surrogate reward. 
(a) IPN trained and \textit{tested on a fixed IC}. 
(b) IPN \textit{tested on random ICs} using policies trained on a fixed IC (from \Cref{fig:burgers_barchart_single_0p9_vertical0p3_surrogate}), policies pretrained on fixed IC and fine-tuned on random ICs, and policies only trained on random ICs. (c/d) Visualization of resulting meshes for Burgers equation with a fixed bump IC at $T=50$.
}
\label{fig:surrogate_reward}
\vspace{-10pt}
\end{figure*}

\section{RESULTS}
\label{sec:results}
\vhalf

We find that the proposed methods achieve performance that is competitive with baselines, outperforming or matching ZZ on 20 out of 24 cases while being competitive with (even sometimes outperforming) the \textit{oracle} TrueError strategy.
More importantly, RL policies generalize well to larger refinement budgets and mesh sizes, and transfer effectively from a static problem to a time-dependent problem.
Almost always, all methods produced the same number of mesh elements at each simulation time step, which means that differences in performance are solely due to differences in refinement strategies.
\Cref{app:runtime} provides information on the cost of training and decision times.
Videos of policies on advection and Burgers can be viewed at \url{https://sites.google.com/view/rl-for-amr}.


\vspace{-10pt}
\subsection{In-distribution}
\label{subsec:results_in-distribution}

\textbf{Static functions} (\Cref{fig:static_train_equal_test}).
RL policies either meet or significantly exceed the performance of ZZ on all function classes.
Notably, both IPN and Graphnet outperform ZZ significantly on \textit{steps} by spending the limited refinement budget only on regions with discontinuities (\Cref{fig:mesh_steps_pg}).
On the smoother function classes such as \textit{bumps} where ZZ is known to perform well, all three policy architectures have comparable performance to ZZ.
Overall, IPN outperforms both Graphnet and Hypernet.
This suggests that capturing higher-order interaction among observations, each of which already contains local neighborhood information, is unnecessary for estimation of static functions as they only have a local domain of influence.
Hypernetwork policies converged to the behavior of making no refinements on at least one out of four independent runs on all classes except \textit{bumps}.
This could be attributed to the inherent difficulty of choosing and training a highly nonlinear model.

\textbf{Advection}
(\Cref{fig:advection_train_equal_test}).
As explained above, we limit the true solutions to smooth functions (\textit{bumps} and \textit{circles}) in the advection case.
Graphnet significantly outperformed ZZ on \textit{circles} and is comparable to TrueError on \textit{bumps}, while IPN is comparable to ZZ on both functions.
Hypernet is comparable to ZZ on \textit{circles} but has high variance across independent runs.
Graphnet's higher performance than other methods indicates that its inductive bias can better represent the local geometric relations between neighboring mesh elements along the circle.

\textbf{Burgers equation}.
In experiments with a single bump function (visualized in \Cref{fig:burgers_single_t50}) as the fixed initial condition (IC) in both train and test, IPN trained with both the exact and surrogate rewards outperformed all baselines (\Cref{fig:burgers_barchart_single_0p9_vertical0p3_surrogate}). 
Surprisingly, the policy trained using the surrogate reward slightly outperformed the policy trained with the exact reward, indicating that: 1) the surrogate reward can be effectively used to train policies without the need for a ground truth solution, as is necessary for general random ICs; 2) because the surrogate reward provides a positive reward whenever a refinement action causes a change in the solution, it effectively acts as an “exploration bonus”, which has been observed in the RL literature to improve performance \citep{tang2017exploration}.
Significantly, \Cref{fig:burgers_pretrainsingle_trainrandom_randomIC_t50} and the linked videos show the learned policy generates a moving refinement ``front'' that anticipates and tracks a moving region that requires refinement.
\subsection{Out-of-distribution (OOD)}

\Cref{fig:burgers_barchart_10x10_steps50_randomIC} shows the performance of IPN RL policies and baselines on Burgers equation with random ICs.
Policies trained on a fixed IC using either the exact or surrogate reward (``Fixed IC, ER'' and ``Fixed IC, SR'')  generalize well to random unseen ICs and still outperform baselines. Moreover, policies that were pretrained with the surrogate reward on the fixed IC for 2k episodes and fine-tuned with the surrogate reward on random ICs for another 2k episodes performed the best (``pretrained, SR'').
Policies trained only on random ICs with the surrogate reward were not as performant, indicating that the training time was not sufficient and that pretraining on a fixed IC is a more efficient approach.

\subsection{Generalization}

The proposed architectures enable a trained policy to generalize well on a test mesh of different sizes and with different budget constraints, because they map from local element features to element selection probabilities. 
As long as local features on the test mesh continue to resemble those seen in training, the trained policy continues to perform even when the global test function was previously unseen.

\textbf{Static$\rightarrow$advection} (\Cref{fig:static_to_advection}).
All static-trained policies demonstrated comparable performance to ZZ and TrueError when tested on \textit{advection-bumps}, while both IPN and Graphnet significantly outperformed ZZ on \textit{advection-circles}.
Surprisingly, static-trained IPN significantly outperforms advection-trained IPN when tested on \textit{advection-circles}, and the static-trained Hypernet does so as well on \textit{advection-bumps}, while static-trained Graphnet maintains comparable performance to its advection-trained counterpart (\Cref{fig:advection_train_equal_test} vs. \Cref{fig:static_to_advection}).
\Cref{fig:advection_test_static} shows that a static-trained policy on \textit{bumps} with $B=10$ correctly refines the region of propagation on \textit{advection-bumps} with $B=50$.

\begin{figure*}[htb!]
\centering
\begin{subfigure}[t]{0.24\linewidth}
    \centering
    \includegraphics[width=1.0\linewidth]{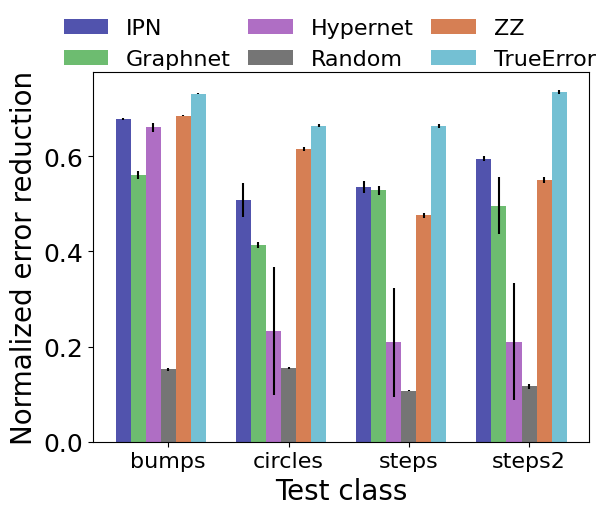}
    \caption{Static $B$=50}
    \label{fig:static_budget_increase}
\end{subfigure}
\hfill
\begin{subfigure}[t]{0.24\linewidth}
    \centering
    \includegraphics[width=1.0\linewidth]{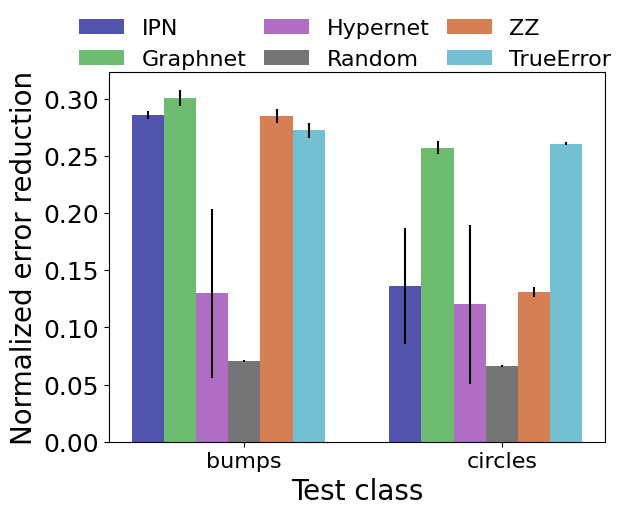}
    \caption{Advection $B$=50}
    \label{fig:advection_budget_increase}
\end{subfigure}
\hfill
\begin{subfigure}[t]{0.24\linewidth}
    \centering
    \includegraphics[width=0.9\linewidth]{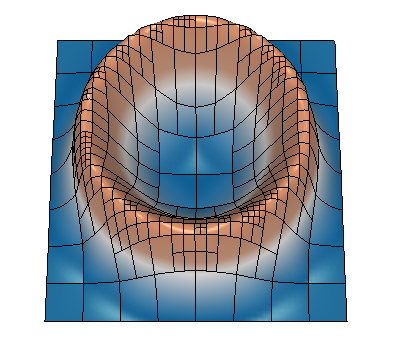}
    \caption{IPN on static circle}
    \label{fig:mesh_circles_pg_train10_test100}
\end{subfigure}
\hfill
\begin{subfigure}[t]{0.24\linewidth}
    \centering
    \includegraphics[width=0.83\linewidth]{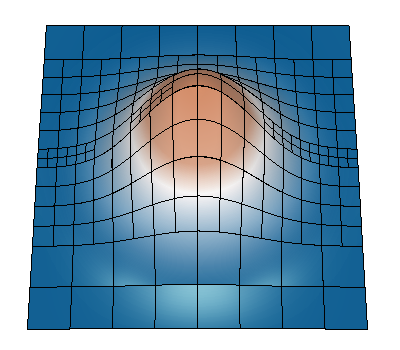}
    \caption{Graphnet on advection}
    \label{fig:advection_bump_pg_graphnet_1_final}
\end{subfigure}
\caption{\textbf{Budget$\uparrow$}. (a-b) Policies trained with budget $B$=10 (static) and $B$=20 (advection) are tested with $B$=50. (c) IPN trained with $B$=10 generalizes to $B$=100. (d) Graphnet trained on advecting bump with $B$=20 generalizes to $B$=50.}
\label{fig:budget_increase}
\end{figure*}

\begin{figure*}[htb!]
\centering
\begin{subfigure}[t]{0.24\linewidth}
    \centering
    \includegraphics[width=1.0\linewidth]{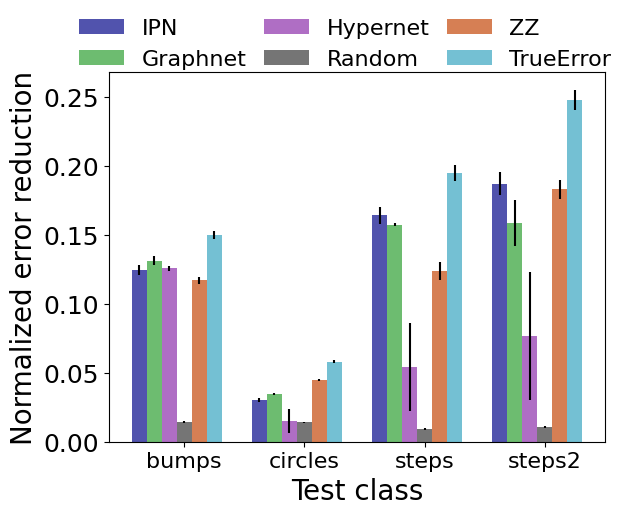}
    \vspace{-.15in}
    \caption{Static 16x16}
    \label{fig:static_density_increase}
\end{subfigure}
\hfill
\begin{subfigure}[t]{0.24\linewidth}
    \centering
    \includegraphics[width=1.0\linewidth]{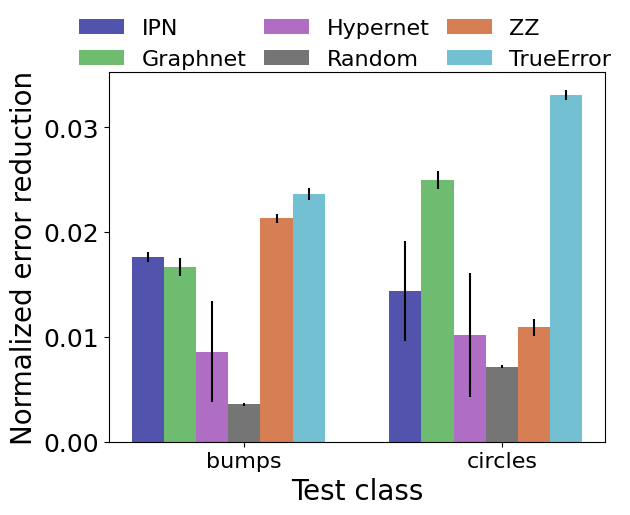}
    \vspace{-.15in}
    \caption{Advection 16x16}
    \label{fig:advection_density_increase}
\end{subfigure}
\hfill
\begin{subfigure}[t]{0.24\linewidth}
    \centering
    \includegraphics[width=0.9\linewidth]{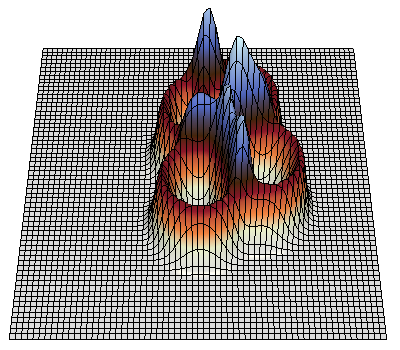}
    \caption{Circles on 64x64 mesh}
    \label{fig:advection_circles_64x64_scale}
\end{subfigure}
\hfill
\begin{subfigure}[t]{0.24\linewidth}
    \centering
    \includegraphics[width=0.95\linewidth]{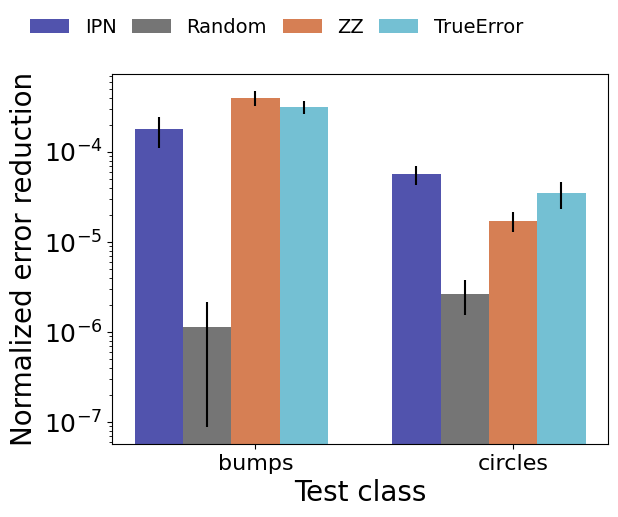}
    \caption{Advection 200x200}
    \label{fig:advection_8to200_steps10_scale_time}
\end{subfigure}
\caption{\textbf{Size$\uparrow$}. (a-b) Policies trained on $8 \times 8$ mesh were tested on $16 \times 16$ mesh. (c-d) Policies trained on $8 \times 8$ mesh were tested on (c) $64 \times 64$ and (d) $200\times 200$ meshes with approximately constant feature-to-mesh length scale ratio.}
\label{fig:density_increase}
\vspace{-10pt}
\end{figure*}

\textbf{Budget$\uparrow$} (\Cref{fig:budget_increase}).
RL policies trained with low refinement budget generalize to test cases with higher budget.
In the static case, comparing \Cref{fig:static_train_equal_test} ($B=10$) with \Cref{fig:static_budget_increase} ($B=50$) shows that the performance of RL policies relative to ZZ is generally preserved by the increase in refinement budget.
\Cref{fig:budget_increase,fig:mesh_static_budget_increase_app} show that an IPN trained with $B=10$ makes qualitatively correct refinement decisions when allowed $B=100$ during test.
In the advection case (\Cref{fig:advection_budget_increase}), Graphnet trained with $B=20$ significantly outperforms both ZZ and TrueError when tested with $B=50$ on \textit{bumps} and comes within the margin of error of TrueError on \textit{circles}.
\Cref{fig:mesh_advection_budget_increase} shows that an IPN trained with $B=20$ correctly allocates a higher budget $B=100$ to the limited region of propagation.

\textbf{Size$\uparrow$} (\Cref{fig:density_increase}).
In the static case, the relative performance of RL policies that were trained with an $8 \times 8$ mesh (\Cref{fig:static_train_equal_test}) is generally preserved when deployed on a $16 \times 16$ mesh (\Cref{fig:static_density_increase}).
All policy architectures outperform ZZ on \textit{bumps}, while IPN and Graphnet still outperform ZZ on \textit{steps}.
IPN and Graphnet were comparable to ZZ on $8 \times 8$ but underperformed on $16 \times 16$ on \textit{circles}.
Nonetheless, \Cref{fig:mesh_circles_ipn_zz} shows that IPN makes qualitatively correct refinements.
On advection, relative performance is preserved on \textit{circles} while IPN and Graphnet deproved slightly on \textit{bumps} (\Cref{fig:advection_train_equal_test} vs. \Cref{fig:advection_density_increase}). Without preserving the solution-to-mesh length scales, the prior tests emulate deploying a policy trained on coarser versions of the simulations.
When tested on $64 \times 64$ and $200\times 200$ meshes (64 and 625-fold increase in number of mesh elements, respectively, e.g. \Cref{fig:advection_circles_64x64_scale}) that preserve the solution-to-mesh length scales, \Cref{fig:advection_8to64_steps20_scale_time,fig:advection_8to200_steps10_scale_time} show that IPN is competitive with baselines on bumps and even outperforms TrueError on circles.
This emulates training a policy on a small subset of a highly-resolved simulation and deploying it on the full simulation.

\vspace{-10pt}
\subsection{Choice of Architecture}
\vspace{-5pt}

For a given test problem, one can choose among the three proposed architectures based on the type of local features that are anticipated to arise.
Such prior knowledge is available for many practical classes of problems that have a long history of instances that were solved by traditional methods.
Comparing overall performance across architectures, \Cref{fig:static_train_equal_test,fig:static_budget_increase,fig:static_density_increase} show that the IPN is the best candidate for problems whose local features are stationary or slowly-propagating (as in the Burgers experiments), whereas \Cref{fig:advection_train_equal_test,fig:advection_budget_increase,fig:advection_density_increase} show that the Graphnet policy is the best candidate for advection problems with smooth features.

\section{LIMITATIONS}
\label{sec:limitations}
\vhalf

Our current approach is limited to refinement of one element per MDP time step.
For practical applications, one can define a ``refinement step'' to consist of multiple MDP time steps, so that multiple elements are refined before the solution is updated and the PDE advances in simulation time.
De-refinement was not included as it requires a multi-objective optimization treatment (error versus cost) that detracts from the main purpose of this work.
Although our results show the RL can outperform greedy baselines, we only conjecture that this is due to anticipatory refinement and leave a deeper investigation to future work.
The scale and complexity of experiments in this work were chosen for agile demonstration of feasibility and generalization, and we hope these promising results serve as a milestone for future application to more complex problems.

\section{CONCLUSION}
\vhalf

We contributed a proof of feasibility for scalable application of reinforcement learning for adaptive mesh refinement.
Our experiments on static and time-dependent problems demonstrate that RL policies can outperform the widely-used ZZ-type error estimator, outperform an oracle based on true error, generalize to different refinement budgets and larger meshes, transfer from static to time-dependent settings, and generalize to more complex problems even when trained without ground truth rewards.
Our finding that RL policies sometimes outperform the true error baseline supports the hypothesis that instantaneous error-based strategies are not optimal due to their inability to refine preemptively.
Future work can extend our methods to include derefinement actions, take a multi-agent perspective, and tackle the unification of $h-$, $p-$, and $r-$ refinement.

\subsubsection*{Acknowledgements}
This work was performed under the auspices of the U.S. Department of Energy by Lawrence Livermore National Laboratory under contract DE-AC52-07NA27344 and the LLNL-LDRD Program with project tracking \#21-SI-001.
LLNL-CONF-820002.


\bibliography{citation}
\bibliographystyle{natbib}

\appendix
\onecolumn



\begin{algorithm}[htb!]
\caption{Graphnet policy forward pass}
\label{alg:graphnet}
\begin{algorithmic}[1]
\FOR{each message-passing round}
\FOR{$k \in \lbrace 1,\dotsc,N^e\rbrace$}
    \STATE $\hat{e}^k \leftarrow \varphi^e(e^k, v^{r^k}, v^{s^k})$ \# Update edge attribute
\ENDFOR
\FOR{$i \in \lbrace 1,\dotsc,N \rbrace$}
    \STATE $\hat{E}^i \coloneqq \lbrace (\hat{e}^k, r^k, s^k)\rbrace_{r^k=i}$ \# Edge set for $v^i$
    \STATE $\bar{e}^i \leftarrow \rho^{e \rightarrow v}(\hat{E}^i)$ \# Aggregation for $v^i$
    \STATE $\hat{v}^i \leftarrow \varphi^v(\bar{e}^i, v^i)$ \# Update vertex attribute
\ENDFOR
\STATE $e^k \leftarrow \hat{e}^k, \forall k \in [N^e]$, $v^i \leftarrow \hat{v}^i, \forall i \in [N]$
\ENDFOR
\STATE $\Rbb \ni x^i \leftarrow \psi(v^i), \forall i \in [N]$
\STATE $\pi(a^i|s)$ is the $i$-th entry of $\text{softmax}(x^1,\dotsc,x^N)$
\end{algorithmic}
\end{algorithm}

\section{Experimental setup}

\subsection{Environment details}
\label{app:environment}

\subsubsection{MFEMCtrl}
\label{app:mfemctrl}

To interface between the MFEM framework and the RL environment, we developed MFEMCtrl, a C++/Python wrapper for the AMR and FEM capabilities in MFEM. MFEMCtrl is used to convert solutions to observations, apply refinement decisions, and calculate errors.


The initial mesh is partitioned into $n_x \times n_y = 8 \times 8$ elements for static and advection experiments and $10 \times 10$ for Burgers equation.
Generalization experiments on larger initial mesh used $n_x \times n_y = 16 \times 16$ or $64 \times 64$. The true solution is projected onto the finite element space by interpolation to the nodes of the Bernstein basis functions. After each refinement action, the solution is projected again onto the refined mesh (for the static case) or integrated in time until the next refinement action (for the time-dependent case). The maximum refinement depth is fixed by the parameter $d_{\text{max}}$ such that the maximally-refined mesh consists of $2^{d_{\text{max}}}n_x \times 2^{d_{\text{max}}}n_y$ elements. $d_{\text{max}}$ was set to 3 for static experiments, whereas $d_{\text{max}}=2$ for advection and $d_{\text{max}}=1$ for Burgers equation due to the time step restrictions imposed by the Courant-Friedrichs-Lewy (CFL) condition of the finest elements.


\textbf{Observation.}\label{app:observation}
The observation consisted of the solution and the depth of each element. Since the gradients of the solution are, by definition, a function of the solution, the observation does not include the gradients as they can be implicitly learned. The solution/depth of each element was observed by interpolating the functions to a local equispaced mesh (\emph{image}) centered around each element, shown by the white box in \Cref{fig:rl_loop}.
Each element's observation is a $l \times w \times c$ tensor where $l = w = l_{\text{element}} + 2l_{\text{context}}$ is the spatial observation window with $l_{\text{element}} = 16$ sampled points inside the element and $l_{\text{context}} = 4$ sampled points in a coordinate direction outside the element.
We chose $c=2$ channels so that estimated function values and element depths are observed, while gradients are omitted since the policy network can in principle estimate gradients from the value channel.
To impose a $1{:}1$ map between each observation and possible action, we append a dummy $o^0$ to state $s$ corresponding to action 0. At most one refinement is allowed per MDP step.

\setlength{\columnsep}{2cm}
\twocolumn

\begin{table}[t]
    \centering
    \caption{Parameterized true solutions}
    \begin{tabular}{ccc}
        \toprule
        & Parameter & [min, max] \\
        \midrule
        \multirow{4}{*}{Bumps (static)}
        & $c_x$ & [0.2, 0.9] \\
        & $c_y$ & [0.2, 0.9] \\
        & $w$ & [0.05, 0.2] \\
        & $n$ & $\lbrace 1,\dotsc,6 \rbrace$ \\
        \midrule
        \multirow{4}{*}{Bumps (advection)}
        & $c_x$ & [0.3, 0.7] \\
        & $c_y$ & [0.3, 0.7] \\
        & $w$ & [0.005, 0.05] \\
        & $n$ & $\lbrace 1,\dotsc,4 \rbrace$ \\
        \midrule
        & $c_x$ & 0.5 \\
        \multirow{1}{*}{Bumps (Burgers)}
        & $c_y$ & 0.5 \\
        \multirow{1}{*}{single IC}
        & $w$ & 0.05 \\
        & $n$ & 1 \\
        \midrule
        & $c_x$ & [0.3, 0.7] \\
        \multirow{1}{*}{Bumps (Burgers)}
        & $c_y$ & [0.3, 0.7] \\
        \multirow{1}{*}{random IC}
        & $w$ & [0.005, 0.05] \\
        & $n$ & $\lbrace 1, \dotsc, 4 \rbrace$\\
        \midrule
        \multirow{5}{*}{Circles (static)}
        & $c_x$ & [0.2, 0.8] \\
        & $c_y$ & [0.2, 0.8] \\
        & $r$ & [0.05, 0.2] \\
        & $w$ & [0.1, 1.0] \\
        & $n$ & $\lbrace 1,\dotsc,6 \rbrace$ \\
        \midrule
        \multirow{5}{*}{Circles (advection)}
        & $c_x$ & [0.3, 0.7] \\
        & $c_y$ & [0.3, 0.7] \\
        & $r$ & [0.05, 0.2] \\
        & $w$ & [0.03, 0.05] \\
        & $n$ & $\lbrace 1,\dotsc,4 \rbrace$ \\
        \midrule
        \multirow{3}{*}{Steps and Steps2}
        & $o$ & [0, 1.0] \\
        & $\theta$ & [0, $\pi/2$] \\
        & $n$ & $\lbrace 1,\dotsc,6 \rbrace$ \\
        \bottomrule
    \end{tabular}
    \label{tab:true_solution}
\end{table}


\subsubsection{Ground truth functions}
\label{app:ground_truth}

Bumps
\begin{align*}
    n &\sim \text{Uniform}[n_{\text{min}}, n_{\text{max}}] \\
    c_{x,i} &\sim \text{Uniform}[c_{x, \text{min}}, c_{x, \text{max}}], \quad i = 1,\dotsc,n \\
    c_{y,i} &\sim \text{Uniform}[c_{y, \text{min}}, c_{y, \text{max}}], \quad i = 1,\dotsc,n \\
    w_i &\sim \text{Uniform}[w_{\text{min}}, w_{\text{max}}], \quad i = 1,\dotsc,n \\
    f(x,y) &= \sum_{i=1}^n \exp\left( -\frac{(x-c_{x,i})^2 + (y-c_{y,i})^2}{w_i} \right)
\end{align*}

Circles
\begin{align*}
    n &\sim \text{Uniform}[n_{\text{min}}, n_{\text{max}}] \\
    c_{x,i}, c_{y,i} &\sim \text{Uniform}[c_{\text{min}}, c_{\text{max}}], \quad i = 1,\dotsc,n \\
    r_i &\sim \text{Uniform}[r_{\text{min}}, r_{\text{max}}], \quad i = 1,\dotsc,n \\
    w_i &\sim \text{Uniform}[w_{\text{min}}, w_{\text{max}}], \quad i = 1,\dotsc,n \\
    f(x,y) &= \sum_{i=1}^n \exp( \\
    & - \frac{(\sqrt{(x-c_{x,i})^2 + (y-c_{y,i})^2}- r_i)^2}{w_i} )
\end{align*}

Steps
\begin{align*}
    n &\sim \text{Uniform}[n_{\text{min}}, n_{\text{max}}] \\
    \theta &\sim \text{Uniform}[\theta_{\text{min}}, \theta_{\text{max}}] \\
    o_i &\sim \text{Uniform}[o_{\text{min}}, o_{\text{max}}], \quad i = 1,\dotsc,n \\
    f(x,y) &= \sum_{i=1}^n 1 + \tanh\left[ 100(o_i - (x + y\tan\theta)) \right]
\end{align*}

Steps2
\begin{align*}
    n &\sim \text{Uniform}[n_{\text{min}}, n_{\text{max}}] \\
    \theta_i &\sim \text{Uniform}[\theta_{\text{min}}, \theta_{\text{max}}], \quad i = 1, \dotsc, n \\
    o_i &\sim \text{Uniform}[o_{\text{min}}, o_{\text{max}}], \quad i = 1,\dotsc,n \\
    s_i &\defeq (x -0.5)\cos\theta_i - (y-0.5)\cos\theta_i \\
    f(x,y) &= \frac{1}{2} \sum_{i=1}^n 1 + \tanh \left[ 100(s_i - o_i) \right]
\end{align*}



\newpage
\onecolumn

\subsubsection{Reward}

Let $u_t$ denote the true solution at time $t$, let $\uhat_t$ denote the estimated solution on the mesh at time $t$.
For a given mesh at time $t-1$, a given time evolution of the true solution from $t-1$ to $t$, and a refinement action $a_t$ (which may be ``do-nothing''), let $\uhat_{t,\text{refine}}$ denote the estimated solution on the mesh at time $t$ that has undergone refinement action $a_t$, and let $\uhat_{t,\text{no-refine}}$ denote the estimated solution on the mesh at time $t$ without that refinement.
We have two reward definitions:
\begin{enumerate}
    \item Delta norm reward with true solution
    \begin{align}
        r_t &\coloneqq (e_{t-1} - e_t) / e_0 \\
        e_t &\coloneqq \lVert u_t - \uhat_t \rVert_2
    \end{align}
    \item Surrogate reward 
    \begin{align}
        r_t &\coloneqq \lVert \uhat_{t,\text{refine}} - \uhat_{t,\text{no-refine}} \rVert_2
    \end{align}
\end{enumerate}
We used the first reward definition for static and advection experiments where analytic true solutions are available, and for Burgers experiments involving a single initial condition and pre-computed reference data that acts as the true solution.
We used the second reward definition for all other Burgers experiments.





\subsection{Implementation}
\label{app:implementation}

We used standard policy gradient \citep{sutton2000policy} for all experiments except for experiments on Burgers equation and generalization of 8x8-trained advection policies to 64x64 test meshes.
We used PPO \citep{schulman2017proximal} for the latter two cases.
We trained for 20k episodes on static problems, 10k episodes on advection problems, 2k episodes on Burgers equation with a single IC, and 4k episodes on Burgers equation with random ICs.
The Burgers experiment with pretraining used 2k episodes on a single IC and a further 2k on random ICs.
Each episode is initialized with refinement budget $B$, where $B=10$ for static problems, $B=20$ for advection, and $B=50$ for Burgers.

\textbf{IPN.}
For efficient computation on a batch of $B$ trajectories, where each trajectory $b$ consists of $T$ environment steps and each step $t_b$ consists of a variable-sized global state $s \in \Rbb^{N_{t_b} \times d}$, we merge the variable dimension with the batch and time dimension to form an input matrix whose dimensions are $[\sum_{b=1}^B \sum_{t=1}^T N_{t_b}, d]$.
The output is reshaped into a ``ragged'' matrix of logits with dimensions $[B\times T, N_{t_b}]$, where the row lengths vary for each batch and time step.
A softmax operation over each row produces the final action probabilities at each step.

\textbf{Graphnet policy.}
The first graph layer is an Independent recurrent block that passes the input node tensors through a convolutional layer followed by a fully-connected layer, to arrive at node embeddings.
This is followed by two recurrent passes through an InterationNetwork \citep{battaglia2016interaction} where fully-connected layers are used for edge and node update functions.
A final InteractionNetwork output layer followed by a global softmax over the graph produces a scalar at each node, which is interpreted as the probability of selecting the corresponding element for refinement.
Except for the input node feature $v^i \in \Rbb^d$ and output node scalar, all internal node (edge) embeddings have the same size, denoted as dim($v$) (dim($e$)).
We fixed $\text{dim}(e)=16$ for both static and advection and tuned $\text{dim}(v)$ (\Cref{tab:hyperparameter}).

\textbf{Hypernet policy.}
We fixed the main network's hidden layer dimension at $h=64$ and tuned the hypernetwork's hidden layer dimension $h_1$ (\Cref{tab:hyperparameter}).

\subsection{Hyperparameters}
\label{app:hyperparameters}

For both static and advection problems, we tuned a subset of all hyperparameters for all methods by the following procedure to handle the large set of policy architectures and ground truth functions.
Chosen values of tuned hyperparameters are given in \Cref{tab:hyperparameter}; all other hyperparameters have the same values for all methods and are listed below.
We conducted tuning in a multi-task setup, where we train a single policy on functions randomly sampled from all ground truth classes, with randomly sampled parameters according to \Cref{app:ground_truth}.
This is done separately on static and advection problems.
The tuning process is coordinate descent where the best parameter from one sweep is used for the next sweep.
We start with exploration decay $\epsilon_{\text{div}} \in \lbrace 100, 500, 1000, 5000 \rbrace$ (a lower bound on exploration was enforced by using behavioral policy $\tilde{\pi}(a_t|s_t) = (1-\epsilon)\pi(a_t|s_t) + \epsilon/N_t$ with $\epsilon$ decaying linearly from $\epsilon_{\text{start}}$ to $\epsilon_{\text{end}}$ by $\epsilon_{\text{div}}$ episodes).
Next we tune the size of hidden layers in the policy network (over $(h_1, h_2) \in \lbrace (128,64),(256,64),(128,128),(256,256) \rbrace$ for IPN, node representation dimension $\text{dim}(v) \in \lbrace 32, 64, 128, 256 \rbrace$ for Graphnet, and $h_1 \in \lbrace 16,32,64,128 \rbrace$ for Hypernet).
Lastly, we tune the learning rate $\alpha \in \lbrace 5\cdot 10^{-5}, 10^{-4}, 5\cdot 10^{-4}, 10^{-3}, 5\cdot 10^{-3} \rbrace$.
For Graphnet and Hypernet, we inherit the best $\epsilon_{\text{div}}$ from IPN because optimal exploration depends in large part on the complexity of the environment, which is the same across all policy architectures.

Separately for the static and advection cases, all three policy architectures have the same values for all other hyperparameters.
These are:
policy gradient batch size 8, initial exploration lower bound $\epsilon_{\text{start}} = 0.5$, final exploration lower bound $\epsilon_{\text{end}} = 0.05$, discount factor $\gamma = 0.99$, convolutional neural network layer with 6 filters of size $(5,5)$ and stride $(2,2)$.

\begin{table*}[ht]
    \centering
    \caption{Hyperparameters for IPN, Graphnet and Hypernet policies on static and advection AMR.}
    \label{tab:hyperparameter}
    \begin{tabular}{lrrrrrr}
        \toprule
        \multicolumn{1}{r}{} &
        \multicolumn{3}{c}{Static} & \multicolumn{3}{c}{Advection}\\
        \cmidrule(r){2-4}
        \cmidrule(r){5-7}
        Parameter & IPN & Graphnet & Hypernet & IPN & Graphnet & Hypernet \\
        \midrule
        $\epsilon_{\text{div}}$ & 500 & 500 & 500 & 100 & 100 & 100 \\
        IPN $(h_1, h_2)$ & (128, 64) & - & - & (256,256) & - & - \\
        Graphnet dim($v$) & - & 64 & - & - & 256 & - \\
        Hypernet $h_1$ & - & - & $128$ & - & - & 32  \\
        $\alpha$ & $10^{-4}$ & $10^{-4}$ & $5\cdot 10^{-5}$ & $10^{-4}$ & $10^{-4}$ & $10^{-4}$ \\
        \bottomrule
    \end{tabular}
\end{table*}

For Burgers experiments and advection experiments on generalization from $8\times 8$ to $64\times 64$ initial mesh sizes, we used a more comprehensive population-based hyperparameter search with successive elimination for all methods.
We start with a batch of $n_{\text{batch}}$ tuples, where each tuple is a combination of hyperparameter values, with each value sampled either log-uniformly from a continuous range or uniformly from a discrete set.
We train independently with each tuple for $n_{\text{episode}}$ episodes, eliminate the lower half of the batch based on their final performance, then initialize the next set of $n_{\text{episode}}$ episodes with the current models for the remaining tuples.
We use the hyperparameters of the last surviving model.
Chosen values are shown in \Cref{tab:hyperparameter_population}.

The hyperparameter ranges are: discount factor in $\lbrace 0.1, 0.5, 0.99 \rbrace$, policy entropy coefficient in $(10^{-3}, 1.0)$, GAE $\lambda$ in $\lbrace 0.85, 0.90, 0.95 \rbrace$, learning rate in $(10^{-5}, 5\cdot 10^{-3}$, PPO $\epsilon$ in $(0.01, 0.5)$, value loss coefficient in $\lbrace 0.1, 0.5, 1.0 \rbrace$, IPN $h_1$ in $\lbrace 128, 256 \rbrace$, and IPN $h_2$ in $\lbrace 64, 128, 256 \rbrace$.

\begin{table*}[ht]
    \centering
    \caption{Hyperparameters for advection $\text{size}\uparrow$ and Burgers experiments}
    \label{tab:hyperparameter_population}
    \begin{tabular}{lrrrr}
        \toprule
        \multicolumn{1}{r}{} &
        \multicolumn{2}{c}{Advection (IPN)} & \multicolumn{2}{c}{Burgers (IPN)}\\
        \cmidrule(r){2-3}
        \cmidrule(r){4-5}
        Parameter & Bumps & Circles & 1 IC & Random IC \\
        \midrule
        Discount $\gamma$ & $0.99$ & $0.99$ & $0.1$ & $0.5$ \\
        Entropy coefficient & $0.0133$ & $0.0689$ & $8.84\cdot 10^{-3}$ & $1.17\cdot 10^{-3}$ \\
        GAE $\lambda$ & $0.95$ & $0.9$ & $0.85$ & $0.85$  \\
        Learning rate & $1.18\cdot 10^{-3}$ & $4.8 \cdot 10^{-3}$  & $1.59\cdot 10^{-3}$ & $2.11\cdot 10^{-4}$ \\
        PPO $\epsilon$ & $0.0113$ & $0.195$ & $0.128$ & $0.169$ \\
        Value loss coefficient & $0.1$ & $0.5$ & $0.5$ & $0.1$ \\
        IPN $h_1$ & $128$ & $256$ & $256$ & $128$ \\
        IPN $h_2$ & $128$ & $128$ & $128$ & $256$ \\
        \bottomrule
    \end{tabular}
\end{table*}

\subsection{Scale of experiments}

The scale and scope of our current experimental setup is comparable with previous work at the intersection of FEM/PDE and machine learning, whose experimental settings are the following:
\begin{enumerate}
    \item 2D Poisson equation with a 64x64 square train mesh \cite{hsieh2018learning}
    \item diffusion PDE on 2D triangular mesh with number of nodes ranging from 1024 to 400k \citep{luz2020learning}
    \item models of simulations with 1k-5k nodes \citep{pfaff2020learning}
    \item Poisson equation on 7x7 mesh with square and sphere domains \citep{alet2019graph}
    \item airfoil with 6648 nodes on a fine mesh and 354 nodes on a coarse mesh \citep{belbute2020combining}
\end{enumerate}

\subsection{Computing infrastructure and runtime}
\label{app:runtime}

Experiments were run on Intel 8-core Xeon E5-2670 CPUs, using one core for each independent policy training session.
Average training time with 20k episodes in the static case was approximately 6 hours for IPN and Hypernet, and 9 hours for Graphnet.
Average training time with 10k episodes in the advection case was approximately 14 hours for IPN and Hypernet, and 18 hours for Graphnet.
In general, the one-shot training time for RL is justified by considering that: 1) given a high-performing trained policy, the time incurred in training is negligible when amortized over all future deployment of that policy; 2) the end-to-end RL approach of optimizing policies directly with experience in simulation may provide a faster path toward new refinement policies that can integrate $h$-, $p$-, and $r$-refinement into a unified strategy, whereas developing new AMR techniques is very human-labor intensive, so the training time of RL policies would be a negligible expense.

\Cref{tab:runtime} shows that test runtimes for all practically-deployable methods are comparable and within the same order of magnitude.
We take the mean over (10 episodes) * (10 steps per episode), which accounts for the impact of variability of different element refinements on the solver time. 
The total differences in runtime between the RL approach and the various baselines manifest only through the differences in the runtime of evaluating the refinement indicator---the RL policy or the baselines---which is reported in \Cref{tab:runtime}.
This is because the time required for computing a solution in between refinement actions is nearly identical between all approaches, as the meshes for all methods are of identical size at each step and there are negligible differences in the computational cost required to solve the PDE on two meshes that differ by only the choice of which elements are refined.
Degrees of freedom (DoF) cost of RL and baselines were the same since all methods refine one element per MDP step in our main experiments.





\begin{table}[H]
\captionsetup{justification=centering}
    \centering
    \caption{ Mean (standard error) time in milliseconds per \\ refinement decision on various initial mesh partitions. }
    \begin{tabular}{lrrrr}
    \toprule
     & $8 \times 8$ & $16 \times 16$ & $24 \times 24$ & $100 \times 100$ \\
    \midrule
    IPN & 3.22 (0.07) & 5.85 (0.05) & 9.64 (0.28) & 158 (5) \\
    Graphnet & 7.74 (0.33) & 13.9 (0.43) & 23.7 (0.19) & 1445 (6) \\
    Hypernet & 8.08 (0.08) & 10.7 (0.05) & 14.3 (0.17) & 151 (9) \\
    ZZ & 1.96 (0.01) & 6.94 (0.01) & 15.5 (0.05) & 134 (3)\\
    \bottomrule
    \end{tabular}
    \label{tab:runtime}
\end{table}

\begin{figure}[t]
\centering
\begin{subfigure}[t]{0.24\linewidth}
    \centering
    \includegraphics[width=1.0\linewidth]{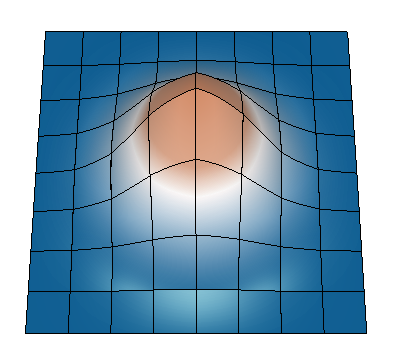}
    \captionsetup{labelformat=empty}
    \caption{$t=0$}
\end{subfigure}
\hfill
\begin{subfigure}[t]{0.24\linewidth}
    \centering
    \includegraphics[width=1.0\linewidth]{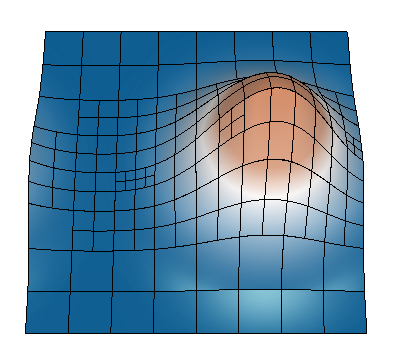}
    \captionsetup{labelformat=empty}
    \caption{$t=32$}
\end{subfigure}
\hfill
\begin{subfigure}[t]{0.24\linewidth}
    \centering
    \includegraphics[width=1.0\linewidth]{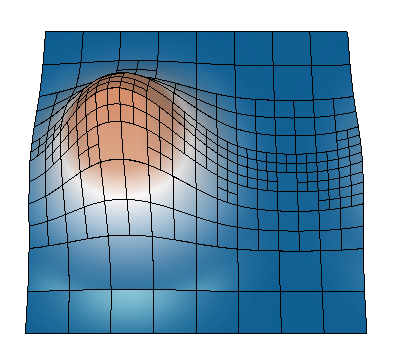}
    \captionsetup{labelformat=empty}
    \caption{$t=68$}
\end{subfigure}
\hfill
\begin{subfigure}[t]{0.24\linewidth}
    \centering
    \includegraphics[width=1.0\linewidth]{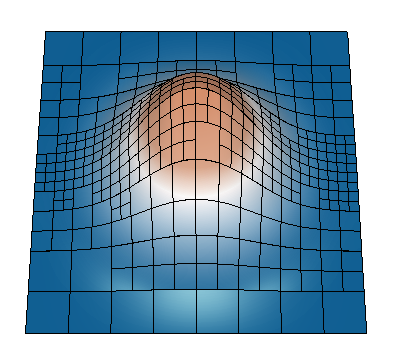}
    \captionsetup{labelformat=empty}
    \caption{$t=100$}
\end{subfigure}
\vspace{-10pt}
\caption{Advection of a bump function. RL policy trained with budget $B=20$ generalizes to $B=100$.}
\label{fig:mesh_advection_budget_increase}
\end{figure}

\begin{figure}[ht]
\centering
\begin{minipage}{0.60\linewidth}
\begin{subfigure}[t]{0.49\linewidth}
    \centering
    \includegraphics[width=1.0\linewidth]{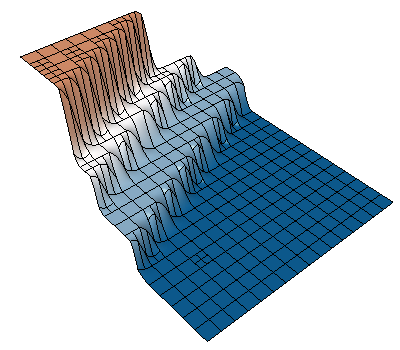}
    \caption{Steps}
    \label{fig:mesh_steps_pg_train10_test100}
\end{subfigure}
\hfill
\begin{subfigure}[t]{0.49\linewidth}
    \centering
    \includegraphics[width=1.0\linewidth]{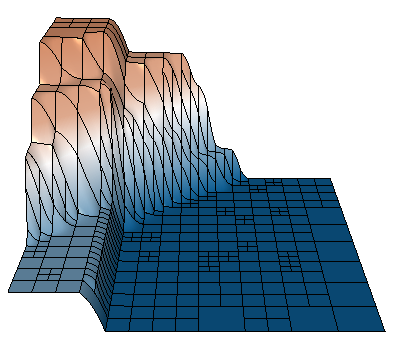}
    \caption{Steps2}
    \label{fig:mesh_steps2_pg_train10_test100}
\end{subfigure}
\caption{Generalization of policies trained with refinement budget $B=10$ to test case with $B=100$.}
\label{fig:mesh_static_budget_increase_app}
\end{minipage}
\hspace{5pt}
\begin{minipage}{0.36\linewidth}
\begin{subfigure}[t]{1.0\linewidth}
    \centering
    \includegraphics[width=1.0\linewidth]{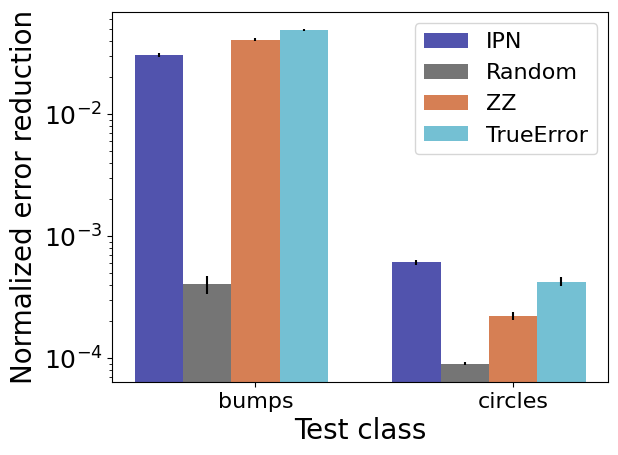}
\end{subfigure}
\vspace{1pt}
\caption{Generalization of policy trained on $8\times 8$ mesh to $64\times64$ mesh.}
\label{fig:advection_8to64_steps20_scale_time}
\end{minipage}
\end{figure}

\begin{figure}[t]
\centering
\begin{minipage}{0.45\linewidth}
\begin{subfigure}[t]{0.49\linewidth}
    \centering
    \includegraphics[width=0.93\linewidth]{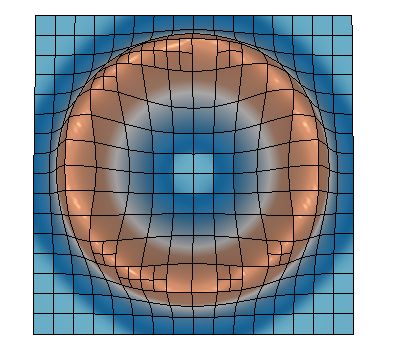}
    \caption{IPN}
\end{subfigure}
\hfill
\begin{subfigure}[t]{0.46\linewidth}
    \centering
    \includegraphics[width=1.0\linewidth]{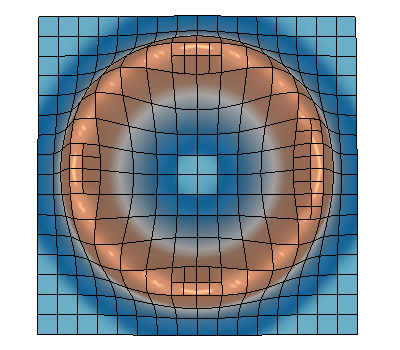}
    \caption{ZZ}
\end{subfigure}
\caption{IPN trained on $8\times 8$ initial mesh underperformed ZZ when tested on $16\times 16$ initial mesh but makes qualitatively correct refinements.}
\label{fig:mesh_circles_ipn_zz}
\end{minipage}
\hspace{5pt}
\begin{minipage}{0.52\linewidth}
    \centering
    \begin{subfigure}[t]{0.49\linewidth}
        \includegraphics[height=0.75\linewidth]{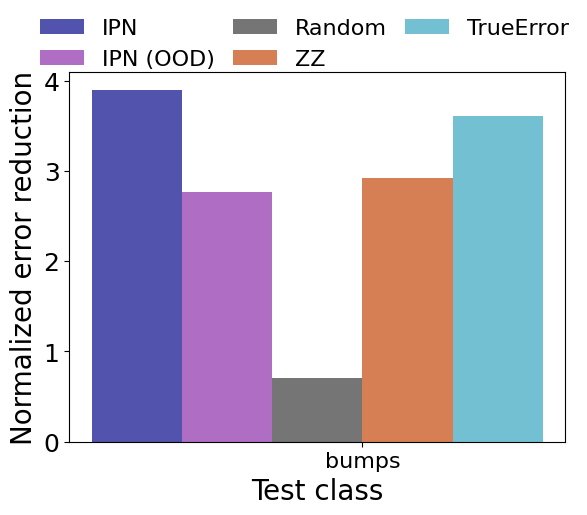}
        \caption{1-bump Burgers}
        \label{fig:burgers_barchart_single_0p9_vertical0p3_surrogate_ood}
    \end{subfigure}
    \hfill
    \begin{subfigure}[t]{0.49\linewidth}
        \includegraphics[width=1.0\linewidth]{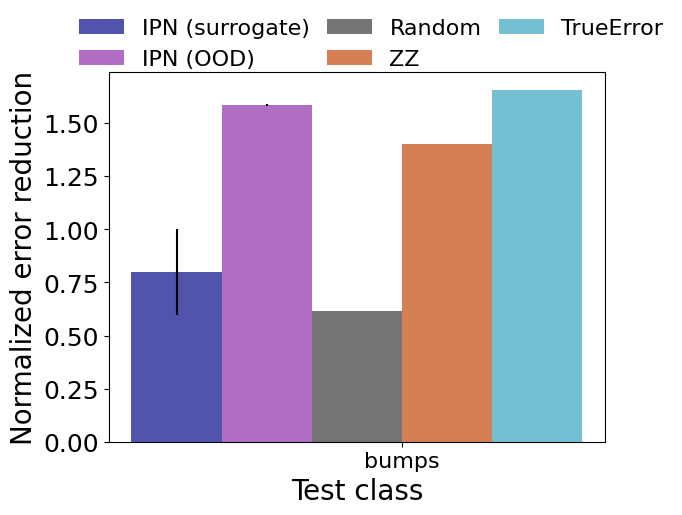}
        \caption{Multi-bumps Burgers}
        \label{fig:burgers_barchart_multi_0p9_vertical0p3_surrogate_ood}
    \end{subfigure}
    \vspace{-5pt}
    \caption{\textbf{OOD}. (a) IPN (OOD) was trained on Burgers with multi-bumps IC and tested on Burgers with a 1-bump IC. (b) IPN (OOD) was trained on 1-bump IC and tested with multi-bumps IC.}
    \label{fig:burgers_ood}
\end{minipage}
\end{figure}

\vspace{-10pt}
\section{Additional results}
\label{app:results}


\begin{wrapfigure}{r}{0.45\textwidth}
    \vspace{-15pt}
    \centering
    \includegraphics[width=0.14\textwidth]{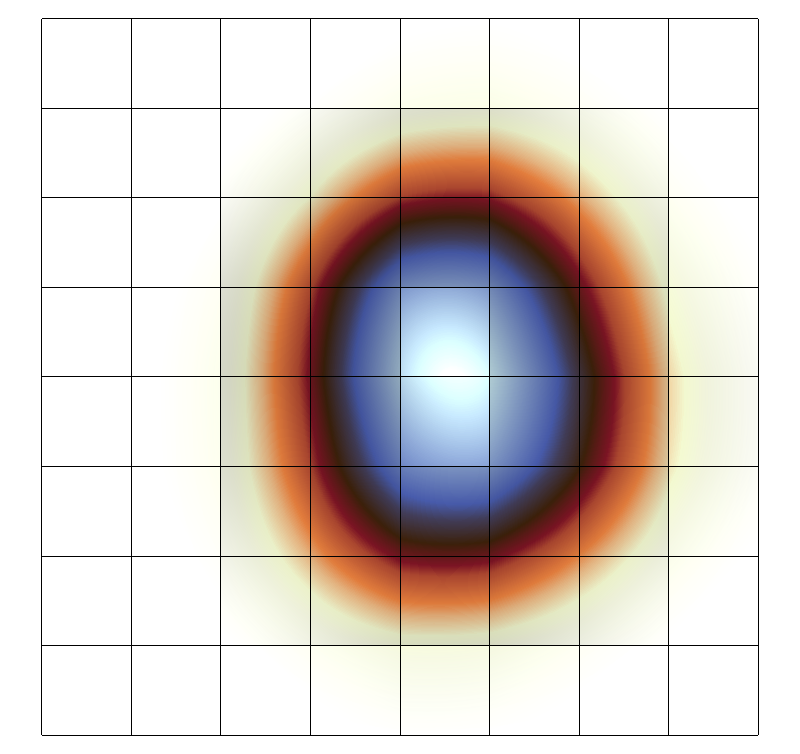}
    \includegraphics[width=0.14\textwidth]{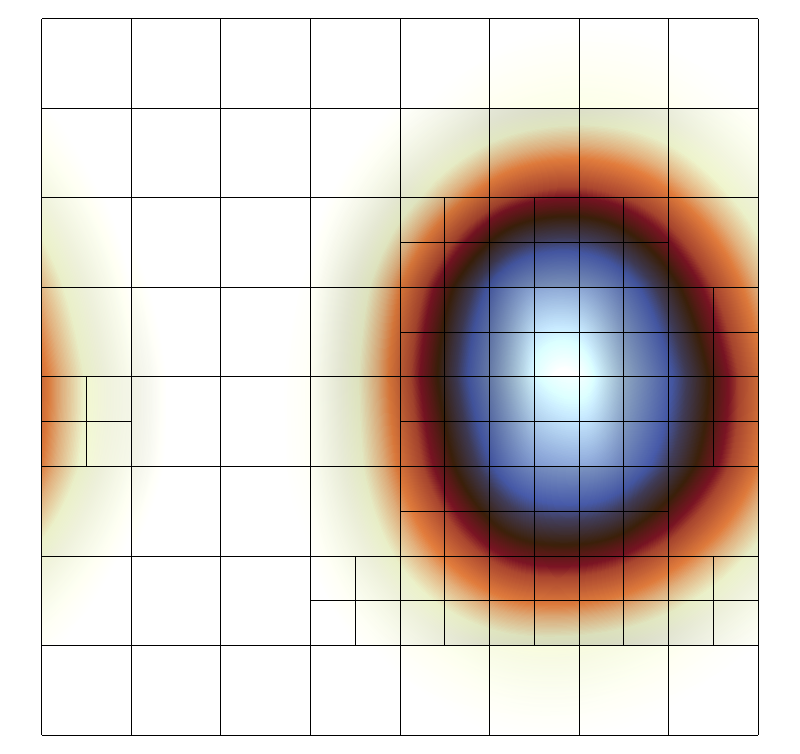}
    \includegraphics[width=0.14\textwidth]{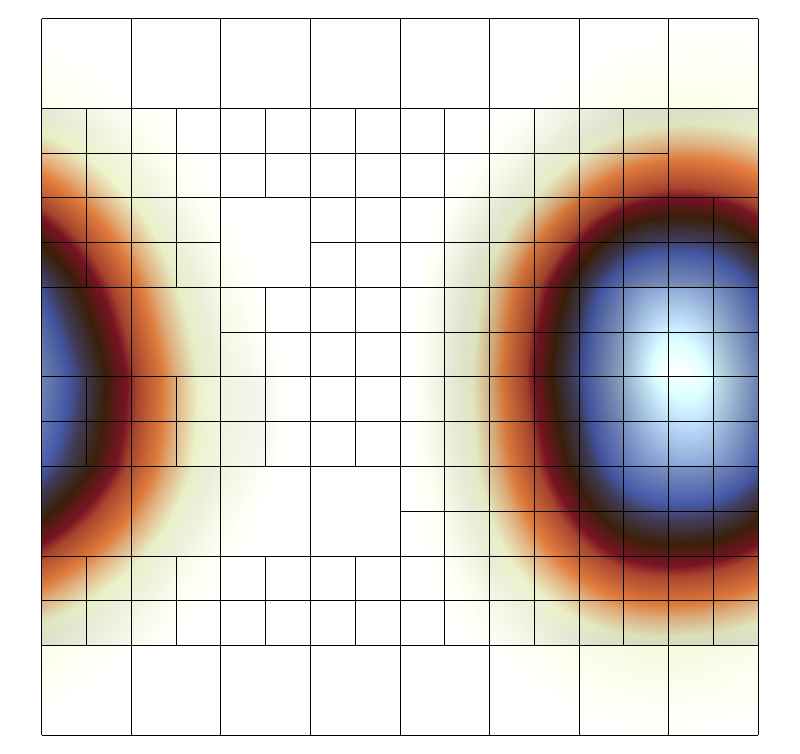}
    \vspace{-10pt}
    \caption{Multiple refinements per solver step.}
    \label{fig:multiple_refinement}
    \vspace{-10pt}
\end{wrapfigure}
For time-dependent problems at training time, only one element is refined for each step that the PDE advances in time.
However, at test time, we can execute the policy multiple times to refine multiple elements before the PDE advances in time.
\Cref{fig:multiple_refinement} shows that the policy makes appropriate choices of 20 elements per step.

\begin{figure}[htb!]
\vspace{-.2in}
\centering
\begin{subfigure}[t]{0.23\linewidth}
    \centering
    \includegraphics[width=1.0\linewidth]{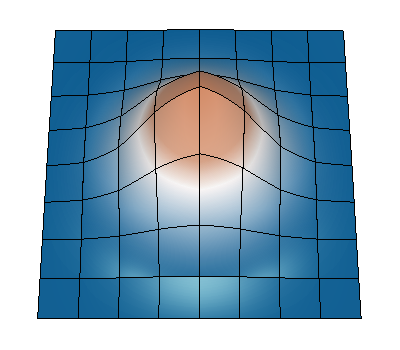}
    \vspace{-.25in}
    \captionsetup{labelformat=empty}
    \caption{$t=0$}
\end{subfigure}
\hfill
\begin{subfigure}[t]{0.23\linewidth}
    \centering
    \includegraphics[width=1.0\linewidth]{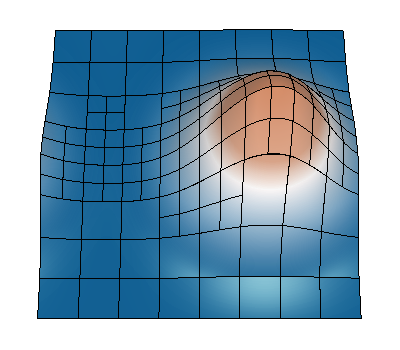}
    \vspace{-.25in}
    \captionsetup{labelformat=empty}
    \caption{$t=22$}
\end{subfigure}
\hfill
\begin{subfigure}[t]{0.23\linewidth}
    \centering
    \includegraphics[width=1.0\linewidth]{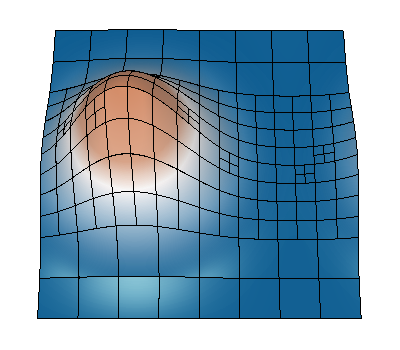}
    \vspace{-.25in}
    \captionsetup{labelformat=empty}
    \caption{$t=38$}
\end{subfigure}
\hfill
\begin{subfigure}[t]{0.23\linewidth}
    \centering
    \includegraphics[width=1.0\linewidth]{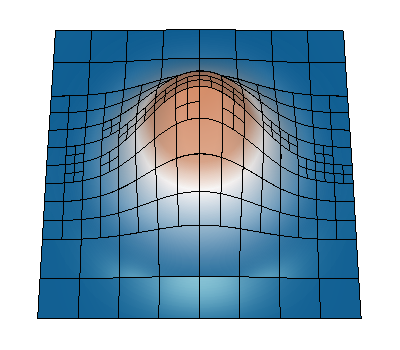}
    \vspace{-.25in}
    \captionsetup{labelformat=empty}
    \caption{$t=50$}
\end{subfigure}
\vspace{-.1in}
\caption{\textbf{Static$\rightarrow$advection} and \textbf{Budget}$\uparrow$: IPN trained on static bumps ($B=10$) transfers to advection ($B=50$).}
\label{fig:advection_test_static}
\end{figure}

\begin{figure*}[t]
\centering
\begin{subfigure}[t]{0.32\linewidth}
    \centering
    \includegraphics[width=1.0\linewidth]{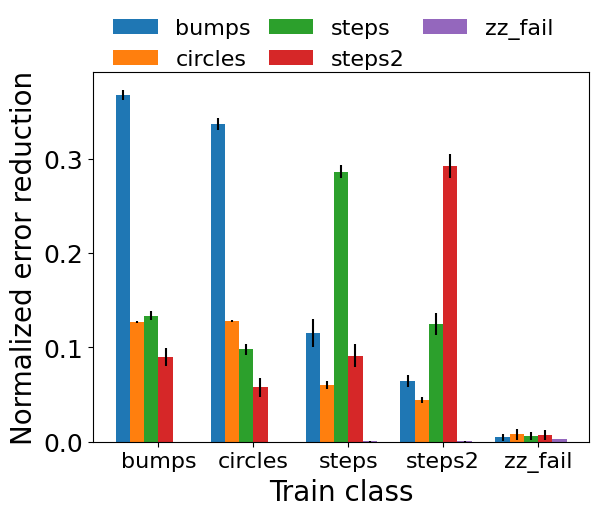}
    \caption{Static IPN}
    \label{fig:static_gen_ipn}
\end{subfigure}
\hfill
\begin{subfigure}[t]{0.32\linewidth}
    \centering
    \includegraphics[width=1.0\linewidth]{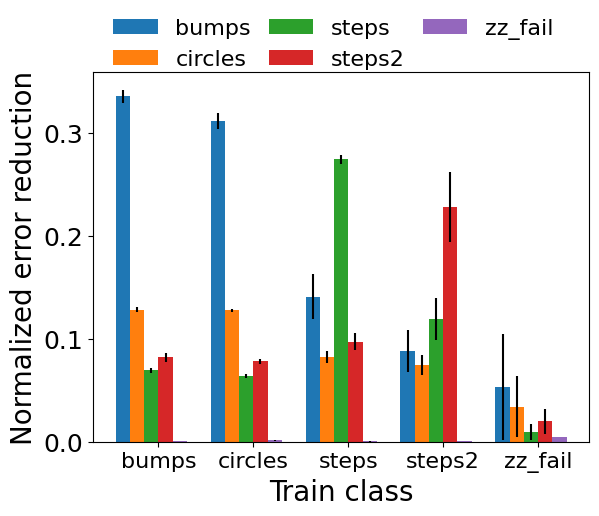}
    \caption{Static Graphnet}
    \label{fig:static_gen_graphnet}
\end{subfigure}
\hfill
\begin{subfigure}[t]{0.32\linewidth}
    \centering
    \includegraphics[width=1.0\linewidth]{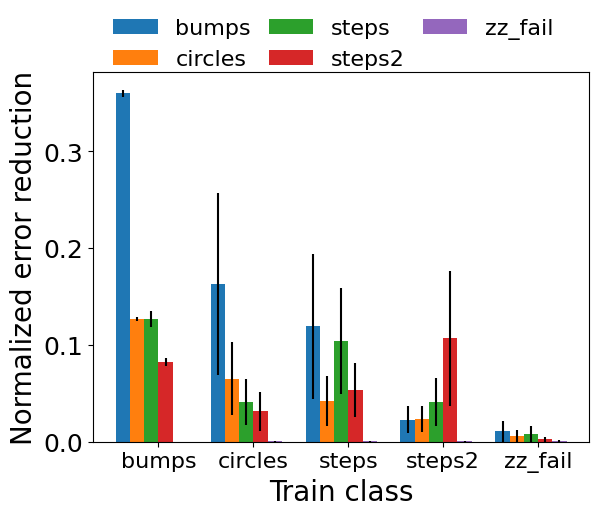}
    \caption{Static Hypernet}
    \label{fig:static_gen_hyper}
\end{subfigure}

\begin{subfigure}[t]{0.32\linewidth}
    \centering
    \includegraphics[width=1.0\linewidth]{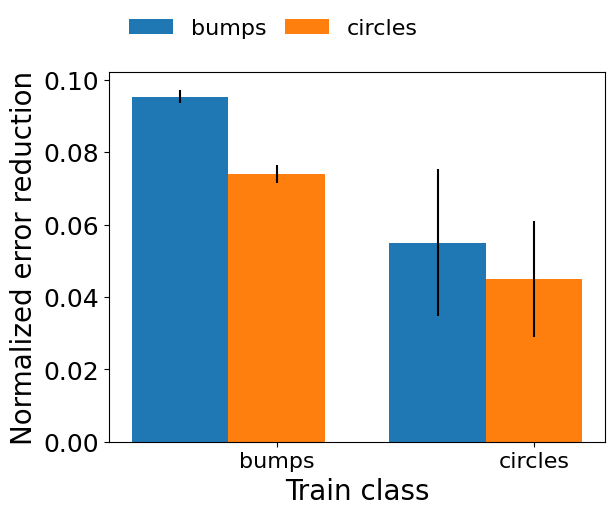}
    \caption{Advection IPN}
    \label{fig:adv_gen_ipn}
\end{subfigure}
\hfill
\begin{subfigure}[t]{0.32\linewidth}
    \centering
    \includegraphics[width=1.0\linewidth]{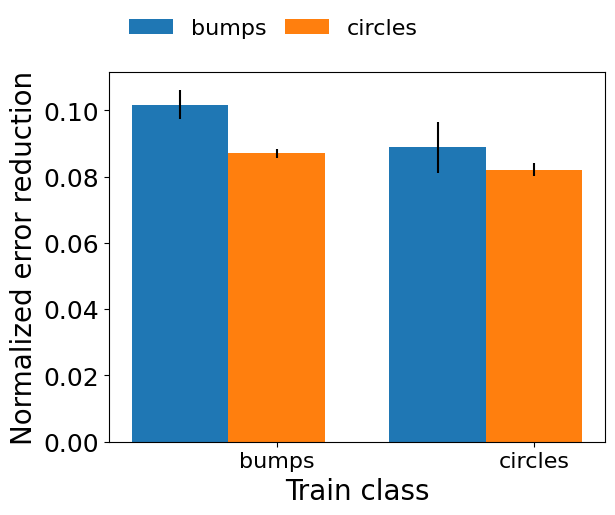}
    \caption{Advection Graphnet}
    \label{fig:adv_gen_graphnet}
\end{subfigure}
\hfill
\begin{subfigure}[t]{0.32\linewidth}
    \centering
    \includegraphics[width=1.0\linewidth]{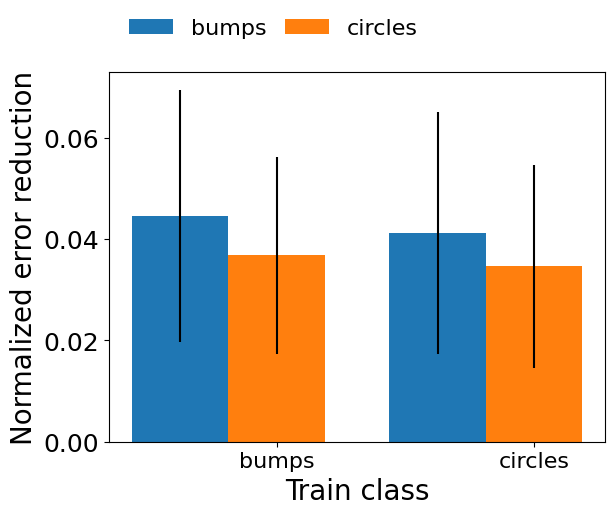}
    \caption{Advection Hypernet}
    \label{fig:adv_gen_hyper}
\end{subfigure}
\vspace{-0.1in}
\caption{All train-test combinations. Normalized error reduction of IPN, Graphnet and Hypernetwork policies on (a-c) Static AMR and (d-f) Advection PDE. Higher values are better. Legend (colors) shows test classes. RL policies were trained and tested on each combination of true solutions. Mean and standard error over four RNG seeds of mean final error over 100 test episodes per method.}
\label{fig:train_test_combinations}
\end{figure*}

\textbf{OOD.}
In the static case (\Cref{fig:static_gen_ipn,fig:static_gen_graphnet,fig:static_gen_hyper}),
IPN policies trained on \textit{circles} transfer well to \textit{bumps} (and vice versa).
Hypernet policies performed poorly overall even in the case of \textbf{in-distribution}, and consequently does not show comparable performance when transferring across function classes.
In the advection case (\Cref{fig:adv_gen_ipn,fig:adv_gen_graphnet,fig:adv_gen_hyper}),
both IPN and Graphnet policies trained on \textit{bumps} significantly outperformed ZZ when tested on \textit{circles} (compare to ZZ in \Cref{fig:advection_train_equal_test}).



\end{document}